\UseRawInputEncoding
\documentclass{article}

\usepackage{microtype}
\usepackage{graphicx}
\usepackage{subcaption}
\usepackage{booktabs} 
\usepackage{hyperref}
\usepackage{longtable}
\usepackage{float}

\usepackage{xurl}

\usepackage[accepted]{icml2026}

\usepackage[T1]{fontenc}
\usepackage{xspace}

\usepackage{amsmath}
\usepackage{amssymb}
\usepackage{amsfonts}
\usepackage{amsthm}
\usepackage{mathtools}

\usepackage[dvipsnames]{xcolor}

\usepackage[table]{xcolor}

\usepackage{multirow}
\usepackage{makecell}
\usepackage{adjustbox}

\usepackage{listings}
\usepackage{upquote}
\usepackage{fvextra}

\usepackage[most]{tcolorbox}
\tcbuselibrary{listings,skins,breakable}

\usepackage[capitalize,noabbrev]{cleveref}

\usepackage{pifont}
\definecolor{g_check}{rgb}{0.0, 0.5, 0.0}
\definecolor{r_check}{rgb}{1.0, 0.0, 0.0}

\usepackage{enumitem}

\theoremstyle{plain}

\theoremstyle{definition}

\theoremstyle{remark}

\usepackage[textsize=tiny]{todonotes}

\newtcblisting{mdblock}{
  listing engine=listings,
  breakable,                 
  enhanced,
  listing only,              
  colback=black!3,
  colframe=black!25,
  boxrule=0.5pt,
  arc=2mm,
  left=4pt,right=4pt,top=4pt,bottom=4pt,
  listing options={
    basicstyle=\ttfamily\small\raggedright, 
    columns=fullflexible,       
    keepspaces=true,            
    showstringspaces=false,
    breaklines=true,            
    breakatwhitespace=true,    
    breakindent=0pt,       
    upquote=true,                
    backgroundcolor=\color{black!3},
  }
}

\usepackage{macros}

\newcommand{\ssa}{\textsc{SpreadsheetArena}\xspace}

\icmltitlerunning{SpreadsheetArena: Decomposing Preference in LLM Spreadsheet Generation}

\begin{document}

\twocolumn[
  \icmltitle{\textsc{SpreadsheetArena}: Decomposing Preference in\\LLM Generation of Spreadsheet Workbooks}



  \icmlsetsymbol{equal}{*}

  \begin{icmlauthorlist}
    \icmlauthor{Srivatsa Kundurthy}{equal,llabs,cornell}
    \icmlauthor{Clara Na}{equal,cmu}\\ 
    \icmlauthor{Michael Handley}{llabs}
    \icmlauthor{Zach Kirshner}{llabs}
    \icmlauthor{Chen Bo Calvin Zhang}{scale}
    \icmlauthor{Manasi Sharma}{scale}
    \icmlauthor{Emma Strubell}{cmu}
    \icmlauthor{John Ling}{llabs}
  \end{icmlauthorlist}

  \icmlaffiliation{llabs}{Longitude Labs Inc., New York, NY}
  \icmlaffiliation{cornell}{Cornell University, Ithaca, NY}
  \icmlaffiliation{cmu}{Carnegie Mellon University, Pittsburgh, PA}
  \icmlaffiliation{scale}{Scale AI, San Francisco, CA}
  
  \icmlcorrespondingauthor{Srivatsa Kundurthy}{srivatsa@meridian.ai}
  \icmlkeywords{Machine Learning, ICML}

  \vskip 0.3in
]


\printAffiliationsAndNotice{\icmlEqualContribution}

\begin{abstract}
We consider the task of end-to-end \textbf{spreadsheet generation}, where language models produce spreadsheet artifacts to satisfy users' explicit and implicit constraints, specified in natural language. 
We introduce \textsc{SpreadsheetArena}, a platform for evaluating models' performance on the task via blind pairwise preference votes of LLM-generated spreadsheet workbooks. As with other complex, open-ended tasks, relevant evaluation criteria can vary greatly across use cases, often in ways that are difficult to formalize. 
Compared to general dialogue or text generation settings, spreadsheet generation presents unique challenges and opportunities: the task output structure is well-defined and multi-dimensional, and there are often complex interactivity and layout considerations. We observe that stylistic, structural, and functional features of preferred spreadsheets vary meaningfully across prompts. Expert evaluations of spreadsheets for finance prompts suggest that even highly ranked models do not reliably produce spreadsheets aligned with domain-specific best practices. We host a live arena and release a dataset of prompts, generated spreadsheets, and preference votes, which we hope will facilitate further study of tasks operating over spreadsheets as a challenging and interesting class of complex, open-ended tasks for LLMs.
\end{abstract}

\vspace{-15pt}
\section{Introduction} \label{sec:introduction}

\begin{figure}[ht!]
    \centering
    \includegraphics[width=\linewidth]{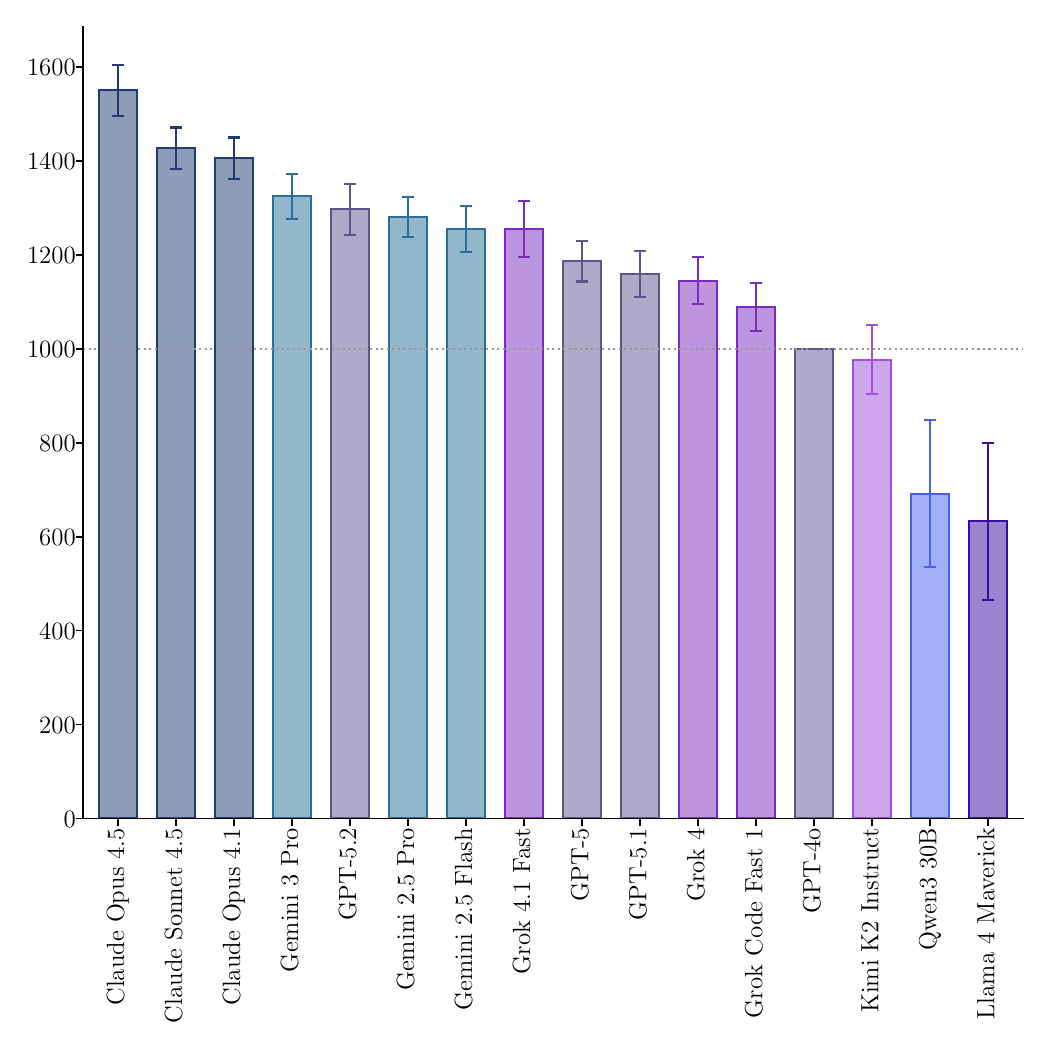}
    \vspace{-20pt}
    \caption{Elo ratings for 16 models ranked in \ssa. Standard Elo scores are anchored on GPT-4o at 1000. Overall, Claude models are often preferred. In \S\ref{sec:analysis} we contextualize these global rankings with observable feature-adjusted scores, category-specific analysis across prompts, characterization of failure modes in dispreferred spreadsheets, and expert evaluations in financial modeling use cases.}
    \label{fig:results_fig1}
    \vspace{-15pt}
\end{figure}

Tasks involving the production or manipulation of structured artifacts are a natural fit for automation with large language models (LLMs), including code generation \citep{chen2021evaluatinglargelanguagemodelscodex, roziere2024codellamaopenfoundation}, table generation and representation \citep{zhang-etal-2024-tablellama, tang-etal-2024-strucbench}, text-to-SQL \citep{yu-etal-2018-spider, lei2025spider2}, and spreadsheet formula generation \citep{chen2021spreadsheetcoder, zhao2024nl2formula}. In some cases, successful task completion can be evaluated through programmatic verification of the outputs. However, many tasks of significant practical value to human users are inherently more open-ended, admitting multiple valid solutions and involving objective and subjective evaluation criteria that may differ across use cases and users. While LLMs are often capable of performing these tasks, evaluation of their capabilities remains a challenge.

\begin{figure*}[ht!]
    \centering
    \includegraphics[width=\linewidth]{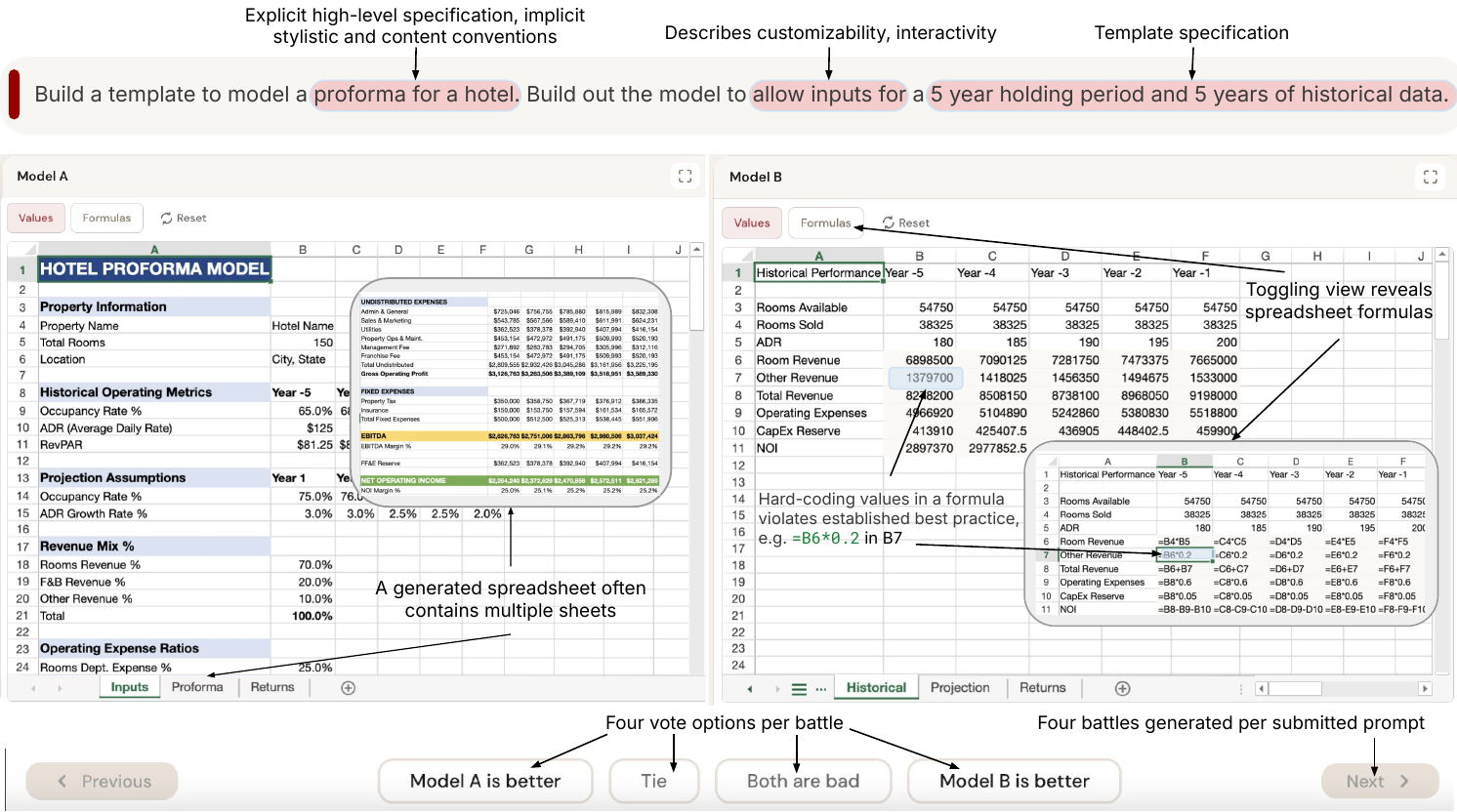} 
    \caption{In \textsc{SpreadsheetArena}, users submit a prompt and are shown four pairwise battles between LLM-generated spreadsheet workbooks. Votes are blind, and users can indicate that one spreadsheet is preferred over the other, or that both are equally satisfactory or unsatisfactory. Workbooks can contain multiple sheets, and sheets often contain a mixture of text, values, and formulas, where cells may contain stylistic formatting (e.g., bold text or a fill color).}
    \label{fig:figure1_interface}
    \vspace{-10pt}
\end{figure*}

We consider end-to-end \textbf{spreadsheet generation} as a task for LLMs, where models are prompted to generate spreadsheet artifacts according to natural language specifications. Use cases for spreadsheet generation span a variety of domains, such as professional finance (e.g., comparing risk across potential investments), academic research (e.g., setting up a statistical significance test given experimental results), and even creative or generative uses (e.g., ``Color in cells to look like Mario''). Criteria for a high-quality spreadsheet workbook output can depend on explicit and implicit contextual factors. One prompt may call for strict adherence to instructions spanning both content and formatting, while another may call for only a template that can be easily updated by the user. Even given a prompt, evaluations may emphasize different criteria, such as correctness of formulas, adherence to domain-specific formatting conventions, or other readability or usability constraints.

Compared to both (1) general open-ended dialogue benchmarks and (2) established tasks involving structured artifact generation, the evaluation of spreadsheet generation presents distinct challenges. Expected outputs are structured artifacts that encode dense, graph-structured dependencies across spreadsheet cells and formulas, exceeding the structural complexity typically seen in open-ended dialogue and even in other commonly studied artifacts such as JSON objects \citep{geng2025jsonschemabench}. Moreover, considerations around user interactivity in spreadsheet workbooks can render errors non-obvious \citep{panko2010taxonomy} and simple execution-based validation insufficient, whereas single-pass execution is common in the evaluation of code generation tasks \citep{chen2021evaluatinglargelanguagemodelscodex, hendrycks2021measuringcodingchallengecompetence}. 

We show that spreadsheet generation is a challenging task presenting opportunities for further study; performant LLMs produce well-formed spreadsheet workbooks with valid formulas more often than not, but practical \textit{functional} utility and adherence to stylistic guidelines, when applicable, are much less reliable. 
Since successful task completion in spreadsheet generation is inherently high-dimensional and context-dependent, human preference evaluation is a critical component of task capability assessment. Towards this, we introduce \textsc{SpreadsheetArena}, a platform for arena-style evaluations of LLM-produced spreadsheet workbooks. 

Additionally, spreadsheet workbook structure enables us to compare voting behaviors with measurable features of winning and losing spreadsheets, such as diversity in formatting, number of filled cells, number of sheets in a workbook, and number of formulas.
Just as response length has been shown to influence text preference evaluations \citep{hu2025explaininglengthbiasllmbased}, we find that certain observable features bear significant influence on model rankings, and that significant features vary across domains. Our findings have implications for post-training with preference data for structured generation tasks, where models must simultaneously satisfy functional, structural, and domain-specific criteria that naive preference data does not uniformly reward. 
In certain domains (professional financial modeling in particular), we additionally contextualize our analyses of preference evaluations with established best practices such as color coding standards, the ``one row, one formula'' rule \citep{FAST2015, WallStreetPrepModeling2020}, and expert evaluations of adherence to finance modeling conventions.

We summarize our core contributions: (1) We introduce \ssa, a platform for evaluating end-to-end spreadsheet generation via blind preference evaluations of spreadsheet workbooks produced by LLMs for user-submitted prompts. The arena is live at \url{https://spreadsheetarena.ai} and contains 4,357 votes over pairwise battles.\footnote{As of January 28th, 2026} (2) We establish stable rankings of 16 LLMs across multiple model families. We show that adjusting for observable features compresses the leaderboard substantially; significant features differ dramatically by domain; and different model families exhibit distinct failure modes. (3) We publicly release a dataset of prompts, spreadsheets, and preference votes for use and further study,\footnote{Dataset: \href{https://huggingface.co/datasets/Longitude-Labs/spreadsheet-arena-release}{https://huggingface.co/datasets/Longitude-Labs/spreadsheet-arena-release}} as well as relevant code.\footnote{Code: \href{https://github.com/Longitude-Labs/spreadsheet-arena}{https://github.com/Longitude-Labs/spreadsheet-arena}}

\paragraph{Conflict of Interest Disclosure} Authors with Longitude Labs Inc. affiliations are employees of the company, which develops and operates the SpreadsheetArena platform introduced in this work and may stand to benefit from its adoption. Longitude Labs also develops a commercial agent for spreadsheet tasks, but this agent is not evaluated in this work, and the platform does not currently accept agentic submissions.

\vspace{-5pt}
\section{Related Work} \label{sec:related_work}
Human preference is central to both post-training and evaluation of large language models. Reinforcement learning from human feedback (RLHF) is a prominent post-training method for aligning model behavior with user intent \citep{christiano2017deep, ouyang2022training} via a reward model \citep{schulman2017ppo}. More recently, \citet{rafailov2023direct} introduced Direct Preference Optimization (DPO), which removes the need for a reward model, but still leverages pairwise comparisons as the supervised learning signal. Preference data also supports evaluation. Preference ``arenas'' collect head-to-head comparisons and aggregate them into rankings via Bradley-Terry or Elo-style estimators \citep{bradleyterry1952rank,coulom2007elo}. \citet{chiang2024chatbotlmarena} introduce LMArena, using blind, community-driven comparisons to rank models. Related efforts such as SEAL Showdown \citep{scaleai2025sealshowdown} emphasize that preference signals can be confounded by factors like verbosity or formatting, motivating analyses that disentangle form from perceived quality \citep{cai2025disentanglinglengthbiaspreference}. While most arenas target conversational settings, the approach is increasingly applied to agentic, tool-using tasks: e.g., the Remote Labor Index (RLI) measures agents on real-world remote-work tasks \citep{mazeika2025remote}. And while rubrics help in conversational settings \citep{lin2024wildbench, arora2025healthbench,akyurek2025prbench}, in agentic settings granular per-project rubrics were often insufficient to capture completion, and for artifacts with hard-to-specify aspects (e.g., design) a deliverable might satisfy rubric elements yet fail professional standards \citep{mazeika2025remote}. Our work evaluates \emph{end-to-end spreadsheet workbook generation}, where user preference reflects functional correctness, formatting, and entangled factors such as vertical-specific style (\cref{sec:analysis}).
\vspace{-10pt}
\paragraph{Finance task evaluation.} A growing body of work trains language models on finance data \citep{wu2023bloomberggptlargelanguagemodel, yang2025fingptopensourcefinanciallarge} and benchmarks them on finance-domain tasks \citep{xie2023pixiulargelanguagemodel, xie2024finbenholisticfinancialbenchmark, zhu2021tatqaquestionansweringbenchmark, chen2022finqadatasetnumericalreasoning}. In parallel, spreadsheet-oriented research addresses table detection and compression \citep{dong2025spreadsheetllmencodingspreadsheetslarge}, formula prediction \citep{chen2021spreadsheetcoder, zhao2024nl2formula}, and manipulation \citep{ma2024spreadsheetbench}. Neither line engages with holistic, preference-based evaluation capturing the multi-dimensional quality considerations—functional correctness, structural organization, and domain-specific conventions (e.g., financial color-coding standards \citep{FAST2015})—that jointly inform real-world financial spreadsheet utility. \ssa addresses this gap, evaluating end-to-end workbook generation through arena-style preference votes complemented by feature decomposition and domain expert evaluation, revealing that general user preferences and expert judgments can substantially diverge.
\vspace{-10pt}

\paragraph{Structured artifacts.} Many prominent tasks require producing or manipulating structured artifacts. Code generation is the most well-studied, given its promise for software and AI automation: LLMs are often trained specifically to generate and reason over code \citep{chen2021evaluatinglargelanguagemodelscodex, roziere2024codellamaopenfoundation}, training corpora \citep{gao2020pile800gbdatasetdiverse, soldaini-etal-2024-dolma, kandpal2025thecommonpile, langlais2026commoncorpus, lambert2025tulu} feature curated code subsets, and code benchmarks are popular for evaluating LLMs \citep{hendrycks2021measuringcodingchallengecompetence, chen2021evaluatinglargelanguagemodelscodex, jiminez2024swebench, deng2025swe}. Tabular and schema-constrained generation have also been studied: \citet{zhang-etal-2024-tablellama} propose TableInstruct, a dataset for instruction fine-tuning, and TableLlama, a model fine-tuned on it. Benchmarks such as StructBench \citep{gu2024structextevalevaluatinglargelanguage}, assessing reasoning over structure-rich text like patient information, and JSONSchemaBench \citep{geng2025jsonschemabench}, evaluating structured output against real-world JSON schema constraints, push this further. Existing spreadsheet benchmarks such as SpreadsheetBench \citep{ma2024spreadsheetbench}, SheetCopilot \citep{li2023sheetcopilot}, and SheetRM \citep{chen2025SheetRMagent} assume manipulation tasks with gold answers that are operationally defined in terms of specific functions, objects (e.g., pivot tables), and cell references. In contrast, \ssa focuses on \emph{end-to-end synthesis of full spreadsheet workbooks} (potentially multi-sheet, with formatting considerations) from tasks often \textit{declaratively} defined by end use case and content; accordingly it uses arena-style preference evaluation to capture holistic utility, complementing purely programmatic metrics.

\section{Background} \label{sec:background}

As noted by \citet{chiang2024chatbotlmarena}, computing rankings from pairwise comparisons is well studied. Following \citet{chiang2024chatbotlmarena} and \citet{scaleai2025sealshowdown}, we apply the Bradley--Terry (BT) model \citep{bradleyterry1952rank} to estimate strength coefficients, from which we derive rankings and Elo-like ratings.

\paragraph{Bradley-Terry model.} The BT model expresses the probability that model $A$ beats $B$ as $P(A \succ B) = \sigma(\theta_A - \theta_B)$, where $\sigma$ is the logistic function and the $\theta$ are strength coefficients. We estimate $\theta$ by maximum likelihood, minimizing the cross-entropy between predicted win probabilities and observed vote outcomes; ordering the resulting coefficients yields rankings that reflect relative win probability.

\paragraph{Elo-like ratings.} Elo and BT parameterize win probabilities as log-odds that are equivalent up to a scaling factor. For interpretability, we convert BT coefficients to Elo-like ratings following \citet{scaleai2025sealshowdown} and \citet{coulom2007elo}. Because the BT model is under-specified \citep{cattelan2012modelsforpairedcomparison}, we anchor a reference model $m_0$ at $\theta_{m_0} = 1000$; we use GPT-4o, the weakest closed model that consistently produces spec-adhering spreadsheets.

\paragraph{Feature-augmented BT.} The standard model attributes performance to a single latent strength and does not capture systematic associations between output features and user preferences. We extend it with feature covariates encoded as pairwise differences, following work on structured BT models with contest-specific effects \citep{cattelan2012modelsforpairedcomparison} and its use for style adjustment in preference arenas \citep{scaleai2025sealshowdown}:
\begin{equation}
P(A \succ B) = \sigma\!\left(\theta_A - \theta_B + \sum_{k=1}^{K} \beta_k (X_{Ak} - X_{Bk})\right), \label{eq:win-probability}
\end{equation}
where $\beta_k$ is the coefficient for feature $k$ and $X_{ik}$ is the mean of feature $k$ over model $i$'s outputs. We use ``control for'' and ``adjust for'' in the regression-adjustment sense: identity parameters $\theta_i$ are estimated conditional on the covariates. Feature-adjusted scores are obtained by subtracting the estimated feature contribution from each output's latent preference score; they are a decomposition of preference signal under the fitted BT model, not an estimate of counterfactual performance under feature manipulation. Conditioning on features shifts $\theta_i$ relative to the vanilla model, and the magnitude of these shifts indicates how much of a ranking difference is attributable to the features (\S\ref{subsec:decomp}).

\section{\textsc{SpreadsheetArena}}
\label{sec:arena_description}
In this section, we introduce \ssa for the evaluation of LLM-produced spreadsheet workbooks. We motivate the arena-style evaluation in the context of the task details and describe our methodology.

\subsection{Task Formulation}
\label{subsec:task_formulation}
In this paper, we study a problem we refer to as spreadsheet generation. In spreadsheet generation, a language model is provided a natural-language text prompt and must produce a spreadsheet artifact. The spreadsheet artifact must be syntactically valid, but beyond syntactic correctness, voting patterns may or may not align with established domain-specific best practices or conventions when applicable.

Spreadsheets occupy a unique position in the landscape of structured artifact generation. Estimates of the global software developer population range from 27 million (professional developers) to 47 million (including students and hobbyists), depending on methodology.\footnote{\href{https://evansdata.com/press/viewRelease.php?pressID=365}{https://evansdata.com/press/viewRelease.php?pressID=365} \href{https://www.slashdata.co/post/global-developer-population-trends-2025-how-many-developers-are-there}{https://slashdata.co/post/global-developer-population-trends-2025-how-many-developers-are-there}} By contrast, Bloomberg estimates that in 2025, there were 500 million paying Excel users,\footnote{\href{https://www.bloomberg.com/features/2025-microsoft-excel-ai-software/}{https://www.bloomberg.com/features/2025-microsoft-excel-ai-software/}} many of whom would not identify as programmers yet routinely build and maintain computation-heavy workbooks. The scale and heterogeneity of spreadsheet users presents distinct evaluation challenges: criteria for a useful, high-quality spreadsheet can depend heavily on explicit and implicit contextual factors that vary across domains, workflows, and user expertise. 

Although spreadsheet generation is a distinct problem with a bounded scope compared to the open-domain chat settings where arena-style evaluations have previously been studied \citep{chiang2024chatbotlmarena, scaleai2025sealshowdown}, user satisfaction signals are similarly relevant for holistic evaluation of generated artifacts. Although the factorization of preference votes to profile the full cross-product of user, prompt, and model characteristics is beyond the scope of this study, we analyze preference votes with spreadsheet and prompt features to conduct targeted investigations of model capabilities and user behaviors across prompt categories.

\subsection{Our Approach}
\label{subsec:approach}
Our task formulation and evaluation methods are agnostic to the spreadsheet synthesis method. In this paper, we explore a setting that assumes a single end-to-end generation of a \textit{serialized representation} of a spreadsheet workbook that is then rendered deterministically. Specifically, models are tasked with generating a JSON representation of a spreadsheet workbook according to the specification described in Appx.~\ref{app:sheetspec}. The schema specifies cell content, sheet structure, and cell style, including, optionally, conditional formatting, over potentially multiple sheets in a workbook.

Alternative approaches to spreadsheet generation may be iterative or agentic; we leave these to future study, and we note that our approach explicitly materializes portable representations of spreadsheet workbooks. These JSON representations are then rendered deterministically in the user's client-side browser via SpreadJS. Where possible, we leveraged support for structured outputs in the model providers' APIs to enforce adherence to our schema. Where not possible at the time of generation, for example for Anthropic models, the schema was appended to the system prompt, also shown in Appx.~\ref{app:sheetspec}.

\subsection{Arena Methodology}
\label{sec:methodology}

\ssa is a platform for pairwise evaluation of LLM-produced spreadsheet workbooks via user vote. Users submit natural language descriptions of their use case or intent, and are shown eight anonymous generated spreadsheet artifacts for each submitted prompt.

As we collect votes, we estimate Bradley-Terry ability parameters \citep{bradleyterry1952rank} for our models. Elo scores \citep{coulom2007elo} are obtained by linearly rescaling the Bradley-Terry parameters, with GPT-4o anchored at 1000. We do not include new models in the leaderboard until they have at least 50 votes.

We initialize \ssa with 436 ``seed'' prompts authored and initially voted on by expert contributors, spanning 6 representative categories of prompts:  Academic \& Research, Corporate Finance \& Financial Planning and Analysis (FP\&A), Creative \& Generative, Operations \& Supply Chain, Professional Finance, and Small/Medium-Sized Business (SMB) \& Personal -- see Appx.~\ref{app:prompt_categories} for details and examples. The taxonomy captures variation in inferrable prompt intent, prompt form and implied context. 
An academic research task might involve finance topics (e.g., regression analysis for computing beta), but the underlying workflow differs fundamentally from professional finance tasks such as indexed stock price returns for a pitch deck.

To classify user-submitted prompts into these categories, we build a prompt categorization pipeline that executes upon prompt submission to auto-categorize prompts on-the-fly. The pipeline uses 1024-dimensional \texttt{Qwen3-Embedding-8B} \citep{qwen3embedding} embeddings of prompts, which are then labeled according to a k-nearest neighbors (k-NN) model fit on the 436 seed prompt embeddings. When a new prompt is submitted, the arena generates pairwise model matches dynamically using Algorithm~\ref{alg:weighted-match-selection}, which prefers models so far seen in relatively fewer battles across the platform. Pairs where at least one model generates an invalid output are discarded and replaced using the same sampling strategy. We detail token usage and execution cost in Appx.~\ref{app:efficiency-analysis}.

\section{Results and Analysis} \label{sec:analysis}

We analyze spreadsheets generated by LLMs in \ssa through arena votes, programmatically extracted spreadsheet features, and expert evaluations. We describe tendencies of different models, variation in use cases, and variation in form and style of winning spreadsheets across domains.

\subsection{General Results}
\label{subsec:basic_results}
We collect a total of 4,357 blind preference votes over pairwise battles between 16 models in \ssa. Table~\ref{tab:model-comparison} contains overall model scores and rankings. Most votes (87.5\%) indicated a preference for one generated spreadsheet over the other. 
Among the remaining battles, 4.0\% were ties (equally as good), and both candidate spreadsheets were judged as unsatisfactory in 8.5\%. In general, prompts with more open-ended use cases (e.g., creative and generative prompts that request drawings or creation of spreadsheet-based puzzles) tend to be more commonly associated with ``both are bad'' votes but are almost nonexistent in others, such as SMB \& Personal use cases. However, for most of our analyses, we use only evaluations where a clear preference of one spreadsheet over the other was indicated.

See Figure~\ref{fig:results_fig1} for a visual ranking of our 16 models' relative performance in the preference arena, and Table~\ref{tab:model_ratings_ci_expanded} in Appx.~\ref{appx:extra_results_ci_etc} for a corresponding table with confidence intervals.

\paragraph{Spreadsheet Preferences vs. Code and Chat Settings} In general chat settings, users prefer longer responses with richer formatting \cite{scaleai2025sealshowdown}. Though there is no spreadsheet feature(s) that is a direct analog to this notion of verbosity or formatting, we do find that significant features corresponding to more text, larger spreadsheets, larger notebooks, more non-empty cells, or more formatting are positively associated with higher win probabilities Table~\ref{tab:feature-effects}. In comparison to code generation in particular, highly rated models in \ssa are often also those that show strong capabilities in coding benchmarks, but high coding benchmark scores are not fully explanatory of \ssa rankings, nor should we assume that spreadsheet generation capability is simply a function of existing tasks.

\paragraph{Evaluation Taxonomies.} We use three complementary evaluation frameworks. (1) We extract a set of 29 \textbf{programmatic features} spanning formula quality, formatting, and structure directly from the spreadsheet artifacts (\S\ref{subsec:decomp}), and analyze their statistical associations with arena preferences. (2) We construct a \textbf{data-driven failure taxonomy} by clustering LLM-generated loss rationales (\S\ref{subsec:loss_buckets}), revealing systematic breakdown patterns not easily captured by scalar features. (3)
We apply an \textbf{expert-designed rubric} grounded in professional finance conventions (\S\ref{subsec:finance_study}), introducing domain-specific normative standards that we find are not well-reflected in crowd preferences. 
Overall, we aim to capture the complexity of spreadsheet generation and its evaluation. Meaningful evaluation requires accounting for heterogeneous preference signals alongside the aggregate performance scores that our global arena rankings provide.

\vspace{-5pt}
\subsection{Preference and Performance Decomposition}
\label{subsec:decomp}

We expand upon methodology from \citet{scaleai2025sealshowdown} and decompose model performance as determined by arena preference votes, by augmenting the vanilla Bradley-Terry model with explanatory feature variables. 
We extract 29 features \textbf{programmatically} from each generated spreadsheet, forming our first evaluation taxonomy. Tab.~\ref{tab:features} in Appx.~\ref{app:spreadsheet_features} details the full set of features. They are distributed across 4 categories that broadly capture spreadsheet quality. \textbf{Formula Quality} features quantify computational correctness and sophistication, including error rates and the use of lookup, conditional, and financial functions; \textbf{Content} features capture the composition of cell types, including text, formulas, and numeric values; \textbf{Formatting} features characterize visual styling such as fills, borders, font treatments, and adherence to professional color-coding conventions; and \textbf{Structure} features describe spatial organization, including sheet dimensions, cell density, and table layouts.

\vspace{-5pt}
\subsubsection{General Feature Effects.}
We fit the augmented Bradley-Terry model in \cref{eq:win-probability} to the parwise \ssa votes, with the full set of 29 spreadsheet features as covariates. \Cref{tab:model-comparison} reports the Elo rating derived from each model's resulting BT coefficient before and after feature adjustments. \Cref{fig:effect_style_controls} in Appx.~\ref{app:spreadsheet_features} visualizes the corresponding shifts.

\begin{table}[htbp]
\caption{Baseline Elo ratings, feature-adjusted Elo ratings, and associated shifts in arena rankings. See \Cref{fig:win_prob_diff} for a visualization. Elo scores are anchored to GPT-4o at 1000. Standard Elo scores correspond to the special case of our Bradley-Terry model without covariates. Feature-adjusted Elo scores are obtained by rescaling the estimated BT model-identity parameters from the covariate-augmented model, evaluated with feature contributions set to zero. We observe substantial compression towards the reference rating in feature-adjusted scores, reflecting reallocation of log-odds mass from model identity to observable output features. While a majority of models experience rank changes, they are limited in magnitude, to only one or two positions.}
\label{tab:model-comparison}
\centering
\scriptsize
\begin{tabular}{@{}lrrrr@{}}
\toprule
\textbf{Model} & \textbf{Baseline Elo} & \textbf{Features Elo} & \textbf{$\Delta$Elo} & \textbf{$\Delta$Rank} \\
\midrule
Claude Opus 4.5   & 1550 & 1333 & $-217$ & 0 \\
Gemini 3 Pro      & 1325 & 1268 & $-56$ & $+2$ \\
Claude Opus 4.1   & 1406 & 1266 & $-140$ & 0 \\
Claude Sonnet 4.5 & 1427 & 1257 & $-170$ & $-2$ \\
Gemini 2.5 Flash  & 1256 & 1225 & $-31$ & $+2$ \\
Gemini 2.5 Pro    & 1279 & 1221 & $-58$ & 0 \\
GPT-5.2           & 1297 & 1175 & $-122$ & $-2$ \\
GPT-5             & 1189 & 1159 & $-30$ & $+1$ \\
Grok 4.1 Fast     & 1255 & 1139 & $-116$ & $-1$ \\
Grok 4            & 1144 & 1132 & $-12$ & $+1$ \\
GPT-5.1           & 1158 & 1125 & $-33$ & $-1$ \\
Grok Code Fast 1  & 1089 & 1108 & $+19$ & 0 \\
Kimi K2 Instruct  & 977 & 1021 & $+44$ & $+1$ \\
GPT-4o            & 1000 & 1000 & 0 & $-1$ \\
Qwen3 30B         & 692 & 849 & $+157$ & 0 \\
Llama 4 Maverick  & 632 & 783 & $+151$ & 0 \\
\bottomrule
\end{tabular}\vspace{-15pt}
\end{table}

\begin{figure}[ht!]
    \centering
    \includegraphics[width=\columnwidth]{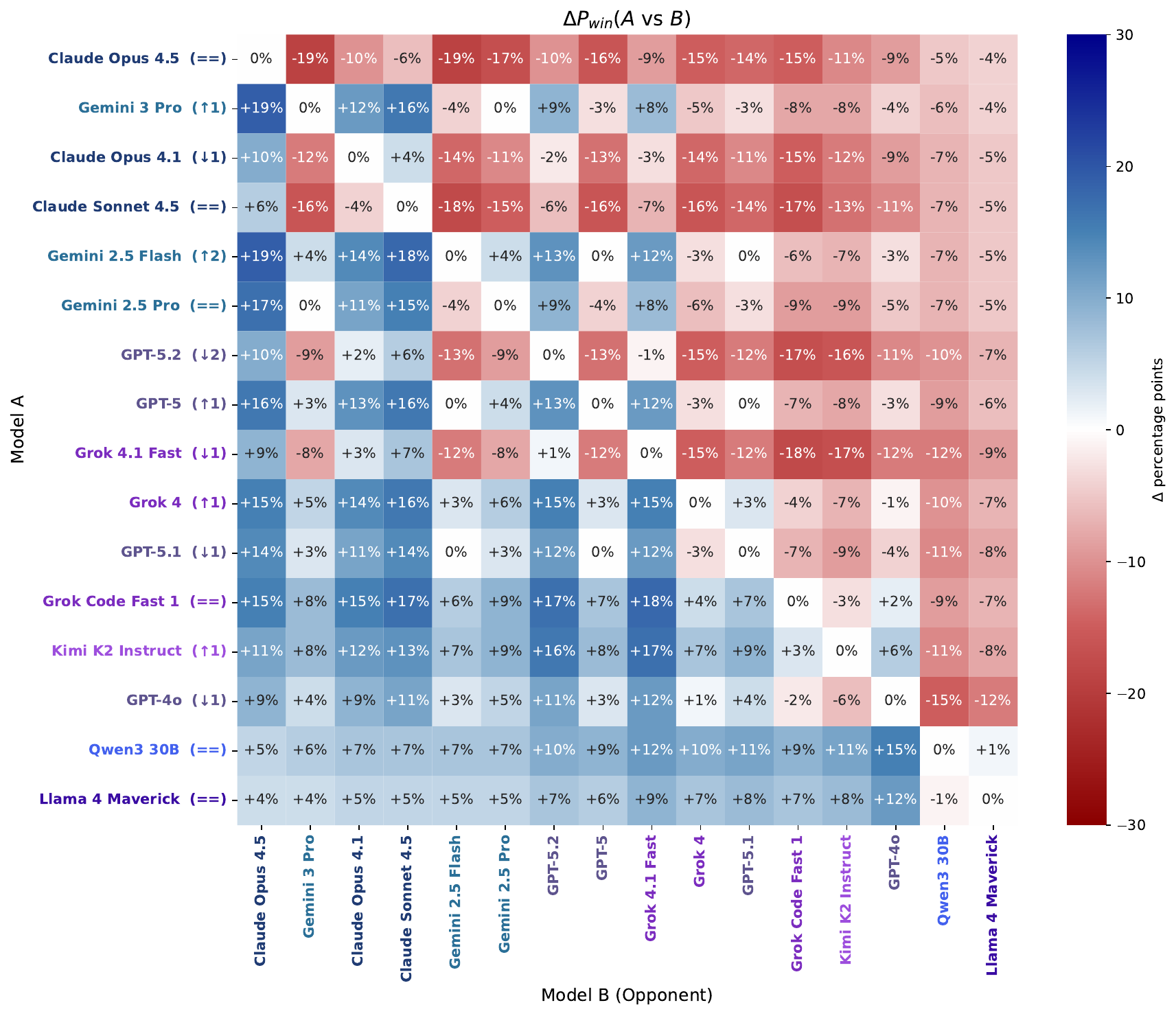} \vspace{-15pt}
    \caption{Pairwise win probability change ($\Delta P_{\text{win}}$) after adjusting for 29 spreadsheet features in the Bradley-Terry model.}
    \label{fig:win_prob_diff}
    \vspace{-20pt}
\end{figure}

\paragraph{Leaderboard Compression.} The most immediate effect of feature controls is a compression of the rating distribution. Claude Opus~4.5 retains the top position but drops 217 Elo points (1550 $\rightarrow$ 1333). Models that underperform in raw rankings show substantial increases in Elo points after controls (Qwen3-30B: 157$\uparrow$, Llama-4-Maverick: 151$\uparrow$). 
The most notable ranking change is Gemini~3~Pro's ascent from 4th to 2nd place, overtaking both Claude Sonnet~4.5 and Claude Opus~4.1. Critically, Gemini~3~Pro experiences only a 56-point Elo decrease, and other Gemini models undergo similarly small ratings shifts when controlling for features, suggesting that Gemini's baseline performance is less confounded by the features we measure; in particular, Claude models seem to have formatting tendencies that happen to align with preference votes. 
Fig.~\ref{fig:win_prob_diff} presents the pairwise win probability changes. We find that feature controls redistribute competitive advantage. Claude Opus~4.5's average win probability against all opponents decreases by 11.2 percentage points on average. 

\paragraph{Which features matter?} Of the 29 features tested, 16 are statistically significant ($p < 0.05$); Tab~\ref{tab:feature-effects} reports their coefficients. The strongest predictors are text density (\texttt{pct\_text}, $+1.56$), background fills ($+1.15$), and numeric content ($+1.02$), features corresponding to explanatory features and formatting.
Formula error rate ($-1.34$) is the strongest negative predictor. The importance of structure is nuanced, with wider layouts being preferred (\texttt{log\_col\_count}, $+0.72$) over fragmented structures such as parallel tables ($-0.21$) and tall aspect ratios ($-0.81$). On the other hand, formula sophistication features do \emph{not} achieve significance: lookup functions ($p = 0.73$), conditionals ($p = 0.28$), and embedded constants ($p = 0.55$) show no reliable effect on win probability. Broadly, complex formula logic does not appear to be rewarded.

\paragraph{Feature significance.} 
Of the 29 features tested, 16 are statistically significant with a standard Benjamini-Hochberg adjustment at $p=0.05$ to control the false discovery rate.\footnote{If we instead apply a Benjamini-Yekutieli correction, a more conservative variant that makes no independence assumptions between features, key conclusions hold, but we lose 5 of our 16 significant features: \texttt{log\_aspect\_ratio}, \texttt{largest\_table\_pct}, \texttt{log\_total\_text\_tokens}, \texttt{log\_distinct\_functions}, \texttt{has\_parallel\_tables}.} Table~\ref{tab:feature-effects} reports their coefficients. The strongest positive associations with win likelihood are text density (\texttt{pct\_text}, $+1.56$), background fills ($+1.15$), and numeric content ($+1.02$), features corresponding to explanatory annotations and formatting. 
Formula error rate ($-1.34$) is the strongest negative association. Effects of structure are more mixed. Wider layouts are preferred (\texttt{log\_col\_count}, $+0.72$) over fragmented structures such as parallel tables ($-0.21$) and tall aspect ratios ($-0.81$). On the other hand, formula sophistication features do \emph{not} achieve significance: lookup functions ($p = 0.73$), conditionals ($p = 0.28$), and embedded constants ($p = 0.55$) show no significant association with win probability. In our BT model, formatting and structural features exhibit stronger associations with preference outcomes than measures of formula complexity. 

\begin{table}[htbp]
\centering
\caption{Statistically significant features. For the full set of features, coefficients, and $p$-values, see \cref{tab:feature-effects-all} in Appx.~\ref{app:spreadsheet_features}.}

\label{tab:feature-effects}
\footnotesize
\begin{tabular}{@{}lr@{}}
\toprule
\textbf{Feature} & \textbf{Coef.} \\
\midrule
\texttt{pct\_text} & $+1.562$ \\
\texttt{compute\_error\_rate} & $-1.338$ \\
\texttt{pct\_fill} & $+1.150$ \\
\texttt{compute\_pct\_numeric} & $+1.020$ \\
\texttt{log\_aspect\_ratio} & $-0.814$ \\
\texttt{log\_col\_count} & $+0.725$ \\
\texttt{pct\_number\_format} & $+0.657$ \\
\texttt{largest\_table\_pct} & $-0.563$ \\
\texttt{has\_border} & $+0.312$ \\
\texttt{log\_num\_blank\_rows} & $-0.248$ \\
\texttt{has\_parallel\_tables} & $-0.214$ \\
\texttt{log\_distinct\_functions} & $-0.211$ \\
\texttt{log\_total\_text\_tokens} & $+0.167$ \\
\texttt{avg\_tables\_per\_sheet} & $+0.104$ \\
\texttt{log\_table\_size\_variance} & $+0.050$ \\
\texttt{num\_single\_cell\_rows} & $-0.027$ \\
\bottomrule
\end{tabular}
\end{table}

\vspace{-5pt}
\subsubsection{Domain Specific Feature Effects}
\label{sec:domain-analysis}

Arena-wide analyses potentially obscure domain-specific preference patterns. We re-estimate our model on each prompt category (merging Professional Finance and Corporate \& FP\&A) and find that feature effects and rankings vary substantially across domains ($p<10^{-68}$ in a likelihood ratio test comparing the pooled augmented BT model against category-stratified models). See Table~\ref{tab:feature_coefficients_category} in Appx.\ref{app:feature_coefs_category_expanded} for full reporting of coefficients in category-specific BT models, and Tables~\ref{tab:academic-rankings} and ~\ref{tab:finance-rankings} in Appx.~\ref{appx:rank-changes} for feature-adjusted rankings in select categories.

We highlight notable coefficient discrepancies and sign reversals across categories. One notable example is \texttt{has\_border}, which is significantly negative for Creative \& Generative ($\beta = -0.87$, $p = .023$) but significantly positive for both Finance ($+0.57$, $.013$) and Operations \& Supply Chain ($+0.71$, $.028$): borders are penalized in generative outputs but rewarded in the more tabular finance and supply-chain settings. 
\texttt{largest\_table\_pct} is significantly negative in Finance ($-1.00$) and Operations \& Supply Chain ($-2.17$) but only non-significantly positive elsewhere, reflecting penalization of table dominance in only some domains. Moreover, \texttt{finance\_color\_convention} is
significant and positive only in Finance ($+1.63$, $.022$).

\subsection{Characterizing Dispreferred Spreadsheets}
\label{subsec:loss_buckets}

\begin{table*}[t]
\centering
\caption{Failure tag rate by model (\% of each model's losses). Models show a high propensity towards presentation failures across the board. Weaker models struggle with prompt alignment and correctness.}
\label{tab:failure_rates}
\footnotesize
\setlength{\tabcolsep}{3pt}
\begin{tabular}{lr|ccccccc}
\hline
 & \textbf{Win} & \textbf{Non-} & \textbf{Spec Non-} & \textbf{Integrity} & \textbf{Numerical} & \textbf{Interpret-} & \textbf{User} & \textbf{Presentation} \\
\textbf{Model} & \textbf{Rate} & \textbf{Functional} & \textbf{compliance} & & \textbf{Computation} & \textbf{ability} & \textbf{Value} & \\
\hline
Claude Opus 4.5        & 83.5\% & 19\% & 18\% & 74\% & 52\% & 52\% & 31\% & 62\% \\
Claude Sonnet 4.5      & 72.4\% & 9\%  & 28\% & 66\% & 45\% & 48\% & 36\% & 57\% \\
Claude Opus 4.1        & 69.2\% & 9\%  & 28\% & 72\% & 46\% & 60\% & 39\% & 81\% \\
Gemini 3 Pro           & 58.3\% & 8\%  & 55\% & 46\% & 36\% & 70\% & 66\% & 85\% \\
GPT-5.2                & 52.7\% & 28\% & 32\% & 57\% & 40\% & 46\% & 51\% & 65\% \\
Gemini 2.5 Pro         & 51.4\% & 15\% & 40\% & 48\% & 33\% & 65\% & 49\% & 88\% \\
Gemini 2.5 Flash       & 51.3\% & 3\%  & 39\% & 24\% & 22\% & 61\% & 59\% & 92\% \\
Grok 4.1 Fast          & 49.5\% & 19\% & 37\% & 53\% & 47\% & 57\% & 63\% & 64\% \\
GPT-5                  & 41.8\% & 12\% & 24\% & 35\% & 20\% & 63\% & 46\% & 88\% \\
GPT-5.1                & 35.2\% & 27\% & 34\% & 57\% & 49\% & 60\% & 48\% & 80\% \\
Grok 4                 & 35.0\% & 23\% & 44\% & 27\% & 11\% & 62\% & 52\% & 96\% \\
Grok Code Fast 1       & 27.1\% & 21\% & 48\% & 60\% & 51\% & 76\% & 60\% & 93\% \\
Kimi K2 Instruct       & 23.7\% & 44\% & 44\% & 63\% & 46\% & 64\% & 54\% & 76\% \\
GPT-4o                 & 20.1\% & 22\% & 68\% & 65\% & 47\% & 55\% & 60\% & 70\% \\
Qwen3 30B              &  9.6\% & 45\% & 77\% & 73\% & 53\% & 83\% & 61\% & 75\% \\
Llama 4 Maverick       &  6.7\% & 20\% & 86\% & 53\% & 35\% & 77\% & 78\% & 87\% \\
\hline
\end{tabular}
\end{table*}

\begin{table}[h]
\centering
\caption{For each of the 7 failure modes in the failure analysis, we compute the Pearson correlation between each model's failure tag rate (\% of that losses tagged with that failure mode) and the model's arena win rate, across all 16 models.}
\label{tab:failure_modes_caught}
\footnotesize
\begin{tabular}{lccc}
\toprule
\textbf{Failure Mode} & \textbf{r} & \textbf{p-value} & \textbf{Arena signal} \\
\midrule
Prompt Miss    & $-0.807$ & $0.0002$ & Caught    \\ 
Low User Value      & $-0.734$ & $0.0012$ & Caught    \\ 
Interpretability      & $-0.661$ & $0.0053$ & Caught    \\ 
Non-functional & $-0.607$ & $0.0127$ & Caught    \\ 
Presentation   & $-0.376$ & $0.1517$ & Ambiguous \\ 
Numerical      & $-0.002$ & $0.9934$ & Missed    \\ 
Integrity      & $+0.061$ & $0.8238$ & Missed    \\ 
\bottomrule
\end{tabular}
\end{table}

To complement our analysis in \S\ref{subsec:decomp} which uses a programmatic feature set, we construct a \textbf{data-driven} failure taxonomy by investigating failure modes of losing candidates. Following \citep{deng2025swe}, we design a taxonomy of tags to support characterization of losing candidate outputs, and subsequently calibrate an LLM judge to apply it to all decisive arena battles. Unlike \citet{deng2025swe}'s error taxonomy that assumes a single ``primary'' failure mode in candidate solutions, however, our categories are explicitly co-occurring diagnostic tags that assume a single losing spreadsheet may exhibit multiple failure modes.

To validate the LLM categorization judge, 5 expert spreadsheet annotators independently labeled a stratified sample of 50 dispreferred spreadsheets, identifying the single most significant failure bucket out of the given taxonomy. The LLM judge's tag set contained the expert-designated primary failure mode in 78\% of cases, indicating strong human alignment with automated review. See Appendix~\ref{appx:failure-modes} for methodological details.

\paragraph{Bucket Definitions.} Each losing spreadsheet is tagged with all categories that contributed to the loss. On average, each losing spreadsheet receives $3.49$ tags, reflecting that spreadsheet failures are typically multi-factorial. *In practice, very few spreadsheets were deemed ``Unjudgeable'' and we merge the label into ``Non-functional.'' 
\begin{description}[leftmargin=0.9cm, labelsep=0.4em, itemsep=0.2ex, parsep=0pt]

\item \textbf{Unjudgeable*}: Cannot be meaningfully evaluated. Empty/truncated or unrelated output.

\item \textbf{Non-functional}: Unusable. Pervasive formula errors block all interpretation of key results. 

\item \textbf{Spec Non-compliance}: Missing core deliverables that the prompt requires. Missing sections, tabs, scenarios, time horizons, or required outputs. 

\item \textbf{Integrity Failure}: Structurally untrustworthy even if surface appearance is plausible. Hardcoded checks, drivers not linked to outputs, or models that do not respond to input changes. 

\item \textbf{Numerical Computation Failure}: Computationally integrated but produces incorrect results. The error is in correctness of the formulas themselves rather than broken linkage or misleading structure. 

\item \textbf{Interpretability Failure}: Hard to follow, teach from, or hand off. Assumptions, calculations, and outputs are not clearly separated. 

\item \textbf{Low User Value}: Correct and readable, but provides no meaningful decision value. 

\item \textbf{Presentation Deficiency}: Inconsistent formatting, nonstandard conventions, or missing visual hierarchy.

\end{description}

\paragraph{Results.}
 Presentation Deficiency is the most pervasive tag, appearing in each model's losses between 57-96\% of the time. Table~\ref{tab:failure_rates} reports the rate at each model's failures are tagged with a given failure mode, demonstrating each model's characteristic failure signature. For example, in 77\% of Qwen3 30B losses, Spec Noncompliance was identified as a contributing factor while 45\% of losing battles were tagged as Non-functional. Similarly, Llama 4 Maverick has an 86\% rate of Spec Non-Compliance. 
 
 Other models exhibit a qualitatively different signature. GPT-5 has fewer errors than the population average in Spec Non-Compliance, Integrity, and Numerical Computation categories, indicating its losses are less likely to stem from missing deliverables or computational errors. Instead, its residual failures are more often associated with Presentation or Interpretability. 

 Notably, the Claude family, though rated favorably in \ssa, shows a distinctive failure profile. Claude Opus 4.5 losses are less often attributed to Spec Non-compliance and Presentation Deficiency relative to the other models  (at 18\% and 62\% respectively),  yet are relatively more often attributed to Integrity and Numerical Computation Errors, at 52\% and 74\% respectively. This suggests Claude's losses are least likely to stem from superficial polish or incomplete outputs. Instead, Claude models' losses are disproportionately related to auditability- and correctness-critical failures that are harder for non-experts to detect but potentially more decisive under expert scrutiny -- this result aligns with the baseline vs. feature-adjusted Elo scores seen in \S\ref{subsec:decomp}.

 Table~\ref{tab:failure_modes_caught} contextualizes LM judge-tagged failure modes with models' arena results. The failure modes with strong negative correlations may be considered as being relevant to user utility: a prompt specification that is missed, a ``correct'' but low-utility spreadsheet, a disorganized structure, or obvious formula errors would be relatively apparent to a user. Failures on numerical correctness and ``best practices'' notions of spreadsheet integrity are caught by experts and an LLM judge but less likely to be punished in arena votes.


\subsection{Finance Domain Expert Evaluation Study} \label{subsec:finance_study}

While arena votes reflect user preferences, they do not directly measure adherence to industry standards. We therefore conduct a blinded expert evaluation of arena-generated spreadsheets from finance-domain prompts, applying an \textbf{expert-designed rubric} to assess whether outputs meet professional modeling standards. Five evaluators with finance-modeling backgrounds (investment banking, private equity) rated 50 spreadsheets from 25 strict win-loss arena battles, blinded to model identity and arena outcome, scoring each on six dimensions (5-point Likert; Table~\ref{tab:finance_eval_dimensions}). In a fully-crossed design, all five experts rated every spreadsheet, yielding 250 evaluations (full protocol in Appx.~\ref{app:finance_evaluation}).

\begin{table}
\caption{Evaluation dimensions for expert annotation of finance-domain spreadsheets}
\label{tab:finance_eval_dimensions}
\begin{center}
\begin{footnotesize}
\begin{tabular}{>{\raggedright\arraybackslash}p{0.3\columnwidth}>{\raggedright\arraybackslash}p{0.6\columnwidth}}
    \toprule
    \textbf{Dimension} & \textbf{Description} \\
    \midrule
    Color Coding, Formatting \& Visual Restraint & Purposeful, consistent formatting that supports readability\\
    \midrule
    Financial Modeling Conventions & Adherence to standard finance modeling norms\\
    \midrule
    Purpose \& Practical Utility & Degree to which the spreadsheet fulfills the prompt and supports decisions\\
    \midrule
    Structure \& Organization & Clear inputs-calculations-outputs flow and auditability\\
    \midrule
    Errors \& Accuracy& Formula correctness and absence of Excel errors\\
    \midrule
    Formula Conventions & Use of best practices for inputs, calculations, and formula design\\
    \bottomrule
\end{tabular}
\end{footnotesize}
\end{center}
\end{table}

\begin{figure}[ht!]
    \centering
    \includegraphics[width=\columnwidth]{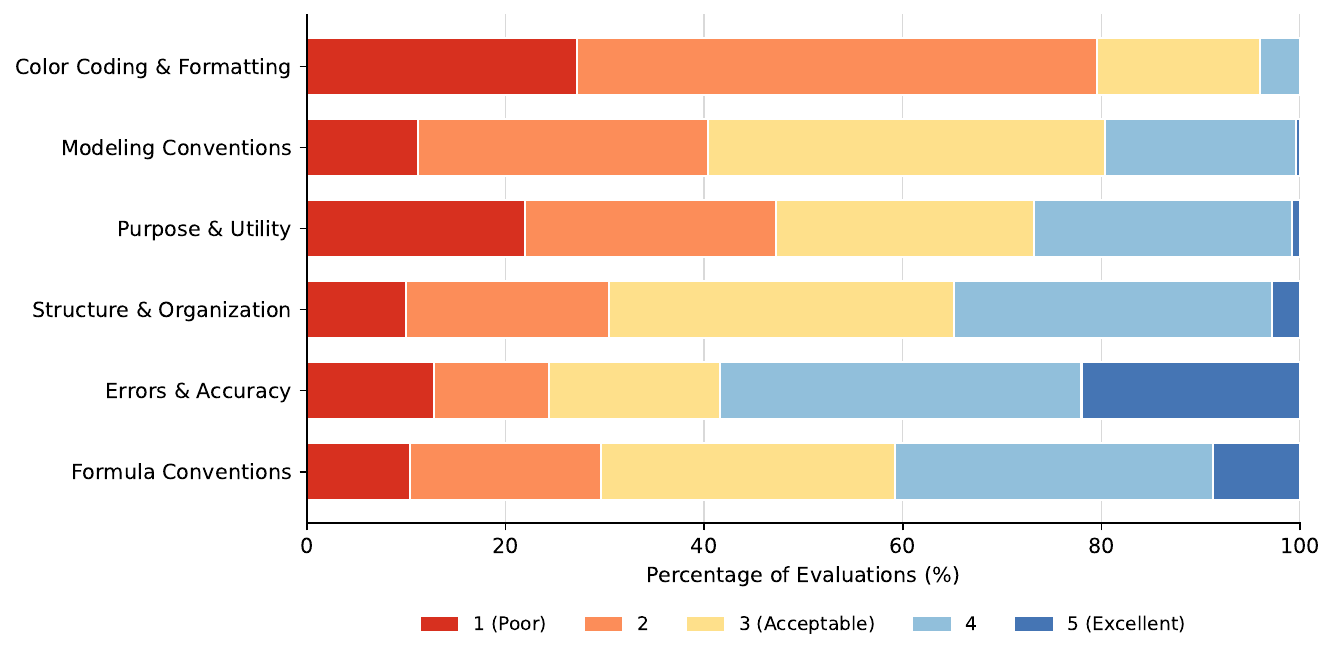}
    \caption{Distribution of expert ratings across six evaluation dimensions for finance-domain spreadsheets ($n=250$ evaluations). Color Coding and Formatting stands out as the weakest dimension, with 79.6\% of evaluations scoring 2 or below.}
    \label{fig:finance_expert_scores}
\end{figure}

\paragraph{Overall performance.} The mean overall rating was 2.86 ($\mathrm{SD}=0.91$), slightly below the midpoint (3 = acceptable): only 25.6\% of evaluations scored $\geq 4$ while 32.0\% scored $\leq 2$, and experts would circulate just 16.1\% of spreadsheets to a client or colleague. Performance was stronger on functional criteria (\emph{Errors \& Accuracy}, $M=3.43$; \emph{Formula Conventions}, $M=3.10$) than on \emph{Modeling Conventions} ($M=2.68$) and \emph{Purpose \& Utility} ($M=2.58$). The largest deficiency was \emph{Color Coding and Formatting} ($M=1.97$), with 79.6\% scoring $\leq 2$: no model consistently followed professional formatting standards (e.g., blue assumptions, black calculations, green cross-sheet links). Per-dimension distributions appear in Fig.~\ref{fig:finance_expert_scores}.

\paragraph{Alignment with arena preferences.} 
Across the 25 battles, expert ratings agreed with the arena outcome in 56.0\% of cases (63.6\% among decisive comparisons), only moderately above chance, and inter-rater reliability was low (Krippendorff's $\alpha = 0.28$--$0.45$; Appx.~\ref{app:finance_evaluation}). This limited alignment suggests that generalized arena preferences may not fully capture finance domain-specific quality requirements, and that LLM spreadsheets show only partial adherence to professional standards.
That being said, our main expert annotation study avoided presenting winning and losing spreadsheets side by side, and so this agreement number can be interpreted as a loose connection between the two evaluations. Instead, we conducted a follow up study asking 5 finance domain experts to each review 25 battles from SpreadsheetArena and pick between File A and File B given the prompt, blinded to model identity and arena outcome. In this formulation, expert-arena alignment is 66.4\%, which is moderate (we expect that they reflect different underlying evaluation criteria) but reasonably higher than the 56.0\% figure.



\clearpage

\section*{Impact Statement}
Large language models have already made outsized impacts on software development. Microsoft CEO Satya Nadella recently reported that as much as 30\% of Microsoft's internal code is now written by AI.\footnote{\url{https://www.cnbc.com/2025/04/29/satya-nadella-says-as-much-as-30percent-of-microsoft-code-is-written-by-ai.html/}} Yet software developers represent a relatively narrow slice of the overall workforce. By contrast, there are an estimated 500 million paying Excel users worldwide, a figure that does not capture the full breadth of spreadsheet usage across other platforms such as Google Sheets.

Spreadsheet usage spans finance professionals, operators, researchers, small business owners, and many other knowledge workers who would not identify as programmers yet routinely build and maintain dynamic, computation-heavy spreadsheet workbooks. If LLMs can achieve for spreadsheets what they have begun to for code, the impact could be substantially broader, reaching hundreds of millions of \emph{end-user developers} across many industries. Our work aims to characterize current capabilities and limitations in LLM-powered spreadsheet generation, providing evaluation infrastructure to support the progress of improved spreadsheet generation systems. 

Additionally, we acknowledge that computational and monetary costs associated with executing spreadsheet generation tasks is non-trivial (see Tables~\ref{tab:token-usage}) and~\ref{tab:api-cost} in Appendix~\ref{app:efficiency-analysis}), though we note that \ssa would cost more if it featured a spreadsheet manipulation arena as well and/or allowed an iterative approach (as opposed to the single pass call we use).

Overall, we argue that spreadsheets are a particularly interesting, understudied domain with potential for significant impact given the hundreds of millions of users of spreadsheet software. Our hope is that this work elucidates current gaps in spreadsheet generation capabilities and inspires further contributions in the space, including both strategies for improving LLM capabilities on the task and evaluations of other related tasks. 

For post-training in particular, our findings suggest that pairwise preference data over structured spreadsheet artifacts does not uniformly reward all dimensions. Notably, formatting features achieve significance while formula sophistication does not, significant features vary across domains, and crowd-sourced preferences agree with expert judgments in finance only modestly. Useful future work may include upstream interventions for improving spreadsheet representation learning, data curation and post-training to improve task-specific generation, exploration of inference algorithms to compare distinct spreadsheet generation paradigms, and scalable evaluations of spreadsheets that are simultaneously grounded in practical, specific user needs.

\section*{Acknowledgements}
We thank Zifan Wang for insightful feedback and discussions during the development of this work.

This material is based upon work supported by the National Science Foundation Graduate Research Fellowship Program under Grant No DGE2140739. Any opinions, findings, and conclusions or recommendations expressed in this material are those of the author(s) and do not necessarily reflect the views of the National Science Foundation.

\bibliography{main}
\bibliographystyle{icml2026}

\newpage
\appendix
\onecolumn

\section{Match Generation Algorithm Details}
In this section, we present the full algorithm for match generation.

\begin{algorithm}[ht]
\caption{\textsc{WeightedMatchEngine}}
\label{alg:weighted-match-selection}
    \begin{algorithmic}[1]
    \REQUIRE Valid models $M$, vote counts $V(\cdot)$
    \ENSURE Set of 4 valid model pairs
    \STATE $P \gets \{(m_i,m_j) \mid m_i,m_j \in M, i<j\}$
    \FORALL{$(m_i,m_j)\in P$}
        \STATE $w_{ij} \gets \left(V(m_i), V(m_j)\right)^{-1/2}$
    \ENDFOR
    \STATE Sort $P$ by decreasing $w_{ij}$
    \STATE $W \gets \emptyset$
    \STATE $k \gets 1$
    \WHILE{$\abs{W} < 4$ and $k \leq \abs{P}$}
        \STATE $(m_i, m_j) \gets P[k]$
        \IF{both $m_i$ and $m_j$ produce valid outputs}
            \STATE $W \gets W \cup \left\{(m_i, m_j)\right\}$
        \ENDIF
        \STATE $k \gets k + 1$
    \ENDWHILE
    \end{algorithmic}
\end{algorithm}

\section{SheetSpec Data Format Specification}
\label{app:sheetspec}

We provide LLMs with a system prompt that calls for an output consisting of only a valid JSON schema representation of a spreadsheet workbook that fulfills the user's request specified in the prompt.

\begin{mdblock}
export const DEFAULT_SYSTEM_PROMPT = """You are a spreadsheet expert.

Return ONLY valid JSON conforming exactly to the provided JSON Schema (SheetSpec@2).
Do not include any explanation, comments, or code fences - output a single JSON object.

All formulas must:
- Use Excel-compatible A1 notation.
- Use commas (,) as argument separators.

Formatting and styling are optional but, if included, must comply with the schema definitions.

Validate that:
- All sheet, column, and cell references used in formulas exist in the output.
- The JSON is syntactically valid and can be parsed directly without modification.""";
\end{mdblock}

For Anthropic models, the \texttt{SheetSpec@2} spec is then appended to this system prompt. For all other models, the structured outputs API option is used to ensure valid schema JSON. A snippet of the full schema is shown below:

\begin{mdblock}
// ...
// ...
// SheetSpec JSON Schema
export const SheetSpecSchema = {
  type: 'object',
  required: ['version', 'sheets'],
  additionalProperties: false,
  properties: {
    version: { type: 'string', const: 'SheetSpec@2' },
    sheets: {
// ...
// ...
export type ConditionalFormatRule =
  | CellIsRule
  | CellIsBetweenRule
  | ExpressionRule
  | ContainsTextRule
  | ColorScaleRule
  | DataBarRule;

export type SheetSpec = {
  version: 'SheetSpec@2';
  sheets: Array<{
    name: string;
    cells: Array<Cell>;
    namedRanges?: Array<{
      name: string;
      ref: string;
    }>;
    conditionalFormats?: Array<ConditionalFormatRule>;
  }>;
  outputs?: Array<{
    name: string;
    sheet: string;
    ref: string;
    metric: 'value' | 'values';
  }>;
  rules?: {
    disallowVolatile?: boolean;
    allowedFunctions?: string[];
  };
};
\end{mdblock}

Cell content can be strings, numerical values, or formulas. Cells can be styled with fills, fonts, borders, and number formatting. Named ranges for formula references are also supported. A substantial subset of Excel's conditional formatting functionality is supported, including value comparisons, custom formulas, color gradients, and data bars. Scale anchors support percentiles and auto-detected min/max values for data-relative formatting.

\clearpage
\section{Model Configurations}
\label{appx:model_configs}

\cref{tab:model_hyperparams} contains model configurations used for our 16 models. 

\begin{table}[h!]
\caption{Model configurations grouped by model provider.} 
\label{tab:model_hyperparams}
\centering
\small
\begin{tabular}{lrr}
\toprule
\textbf{Model Name} & \textbf{Temp} & \textbf{Tokens} \\
\midrule
\textbf{OpenAI (GPT)} & & \\
GPT-5 & default & 60,000 \\
GPT-5.2 & 0.7 & 128,000 \\
GPT-5.1 & 0.7 & 128,000 \\
GPT-4o & default & 16,384 \\
\midrule
\textbf{Anthropic (Claude)} & & \\
Claude Opus 4.5 & 0.7 & 64,000 \\
Claude Opus 4.1 & 0.7 & 32,000 \\
Claude Sonnet 4.5 & 0.7 & 60,000 \\
\midrule
\textbf{Google (Gemini)} & & \\
Gemini 3 Pro & 0.7 & 64,000 \\
Gemini 2.5 Pro & 0.7 & 60,000 \\
Gemini 2.5 Flash & 0.7 & 60,000 \\
\midrule
\textbf{xAI (Grok)} & & \\
Grok 4.1 Fast & default & 2,000,000 \\
Grok Code Fast 1 & 0.7 & 200,000 \\
Grok 4 & 0.7 & 60,000 \\
\midrule
\textbf{Meta (Llama)} & & \\
Llama 4 Maverick & 0.7 & 1,000,000 \\
\midrule
\textbf{Alibaba (Qwen)} & & \\
Qwen3 30B & 0.7 & 128,000 \\
\midrule
\textbf{Moonshot (Kimi)} & & \\
Kimi K2 Instruct & 0.7 & 256,000 \\
\bottomrule
\end{tabular}
\end{table}

\section{Prompt Category Details}
\label{app:prompt_categories}

\subsection{Category Prompt Distribution}

\begin{table}[th!]
  \caption{Category distribution of the $n=1033$ prompts, split by source (Seed vs.\ Arena). Counts and column-wise percentages.}
  \label{tab:prompt_categories}
  \vskip 0.15in
  \begin{center}
  \begin{tabular}{@{}lcccccc@{}}
    \toprule
    & \multicolumn{2}{c}{Total} & \multicolumn{2}{c}{Seed} & \multicolumn{2}{c}{Arena} \\
    \cmidrule(lr){2-3}\cmidrule(lr){4-5}\cmidrule(lr){6-7}
    Category & $n$ & \% & $n$ & \% & $n$ & \% \\
    \midrule
    Academic \& Research        & 56   & 5.4\%  & 36  & 8.3\%  & 20  & 3.4\%  \\
    Corporate Finance \& FP\&A  & 222  & 21.5\% & 97  & 22.2\% & 125 & 20.9\% \\
    Creative \& Generative      & 136  & 13.2\% & 71  & 16.3\% & 65  & 10.9\% \\
    Operations \& Supply Chain  & 116  & 11.2\% & 71  & 16.3\% & 45  & 7.5\%  \\
    Professional Finance        & 261  & 25.3\% & 89  & 20.4\% & 172 & 28.8\% \\
    SMB \& Personal             & 242  & 23.4\% & 72  & 16.5\% & 170 & 28.5\% \\
    \midrule
    \textbf{Total}              & \textbf{1033} & & \textbf{436} & & \textbf{597} & \\
    \bottomrule
  \end{tabular}
  \end{center}
\end{table}

\subsection{Category Prompt Examples}
\label{app:category_examples}

\begin{small}
\begin{longtable}{p{3.2cm} p{11.5cm}}
\toprule
\textbf{Category} & \textbf{Task Description} \\
\midrule
\endfirsthead
\toprule
\textbf{Category} & \textbf{Task Description} \\
\midrule
\endhead
\midrule
\multicolumn{2}{r}{\textit{Continued on next page}} \\
\endfoot
\bottomrule
\endlastfoot

Academic \& Research & 
Create a spreadsheet to perform a difference-in-differences analysis for a policy intervention study. Set up two groups (treatment and control) with pre-intervention data for 2019--2020 and post-intervention data for 2021--2022. Include 8 observations per group with outcome variables showing baseline values around 50 for both groups, then treatment group increasing to around 65 post-intervention while control stays at 52. Calculate the difference-in-differences estimator, parallel trends assumption check, and standard errors. Include a simple visualization comparing the trends. \\
\midrule

Corporate Finance \& FP\&A & 
Build a pricing and margin sensitivity model for a software business to help an entrepreneur understand how pricing changes impact profitability. Assume the business has 1,000 active customers, with monthly churn of 4\% and 100 new customers added per month. Model three pricing scenarios: \$20, \$35, and \$50 per month. Gross margin is 75\% at \$20, 80\% at \$35, and 85\% at \$50. Fixed operating costs are \$40,000 per month. Show monthly revenue, gross profit, operating profit, and break-even point under each pricing scenario, and clearly compare outcomes side-by-side in a sensitivity table. Build with months across columns. \\
\midrule

Creative \& Generative & 
Create a playable Checkers game in a spreadsheet. The 8$\times$8 board should use shaded dark squares (playable) and locked light squares. Pieces use symbols: red = ``r'', black = ``b'', kings = ``R''/``B''. Implement click-based movement with alternating turns, legal diagonal moves only, mandatory jump captures with multi-jump enforcement, and automatic king promotion. Include illegal move prevention, turn indicator, captured piece counts, win/loss/draw detection, conditional formatting for valid moves and captures, and a ``New Game'' reset button. 
\\
\midrule

Operations \& Supply Chain & 
Create a centralized hiring tracker that logs incoming resumes and tracks candidates through each stage of the hiring process. Include applicant details, role applied for, screening status, interview stage, interview feedback, decision outcomes, and timelines. Add automatic status updates, time-to-hire metrics, funnel conversion rates, and visual summaries showing pipeline health and bottlenecks. Design as a reusable template with customizable stages, roles, and evaluation criteria. \\
\midrule

Professional Finance & 
Build a fully integrated, institutional-quality leveraged buyout (LBO) model for a multi-segment operating company with three business segments: one cyclical, one subscription-based recurring revenue, and one capital-intensive legacy segment in decline. Finance the acquisition with a layered capital structure: revolver with cash sweep, Term Loan B with mandatory amortization, PIK toggle mezzanine tranche, seller notes with contingent interest, and rolled management equity with dilution mechanics. Project detailed operating assumptions per segment (revenue drivers, pricing vs.\ volume, gross margin bridges, SG\&A leverage, maintenance vs.\ growth capex, working capital as function of revenue), consolidate into fully linked financial statements. Include transaction/financing fees, OID, deferred financing costs, goodwill/intangibles amortization, quarterly covenant testing (leverage, coverage) with breach triggers, excess cash flow sweeps, and PIK capitalization. Model scenario-based exits with sponsor IRR, MOIC, and cash-on-cash returns. Include sensitivity tables for leverage, entry/exit multiples, operating performance, and interest rates. \\
\midrule

SMB \& Personal & 
Create a weekly food tracker for calorie input from food and exercise output. Include an input area for current weight and target weight. Track calories in and calories out to facilitate weight loss monitoring. \\
\end{longtable}
\end{small}

\subsection{Category Spreadsheet Examples}
\label{app:category_spreadsheet_examples}

\subsubsection{Academic \& Research}

\begin{figure}[H]
    \centering
    \includegraphics[width=0.8\textwidth]{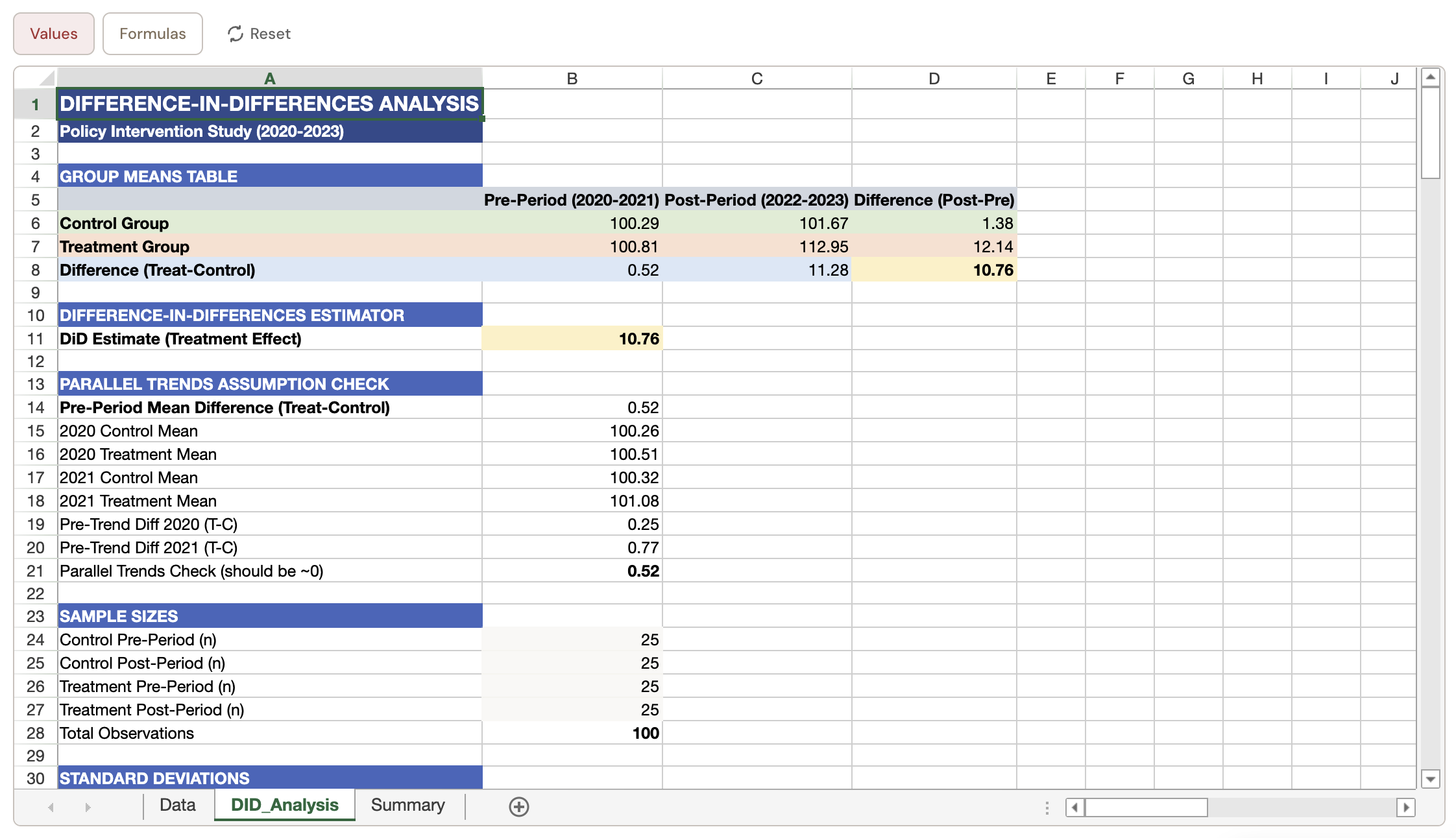}
    \caption{A model response to the ``Academic \& Research'' prompt in Appx.~\ref{app:category_examples}.}
\end{figure}

\subsubsection{Corporate Finance \& FP\&A}

\begin{figure}[H]
    \centering
    \includegraphics[width=0.8\textwidth]{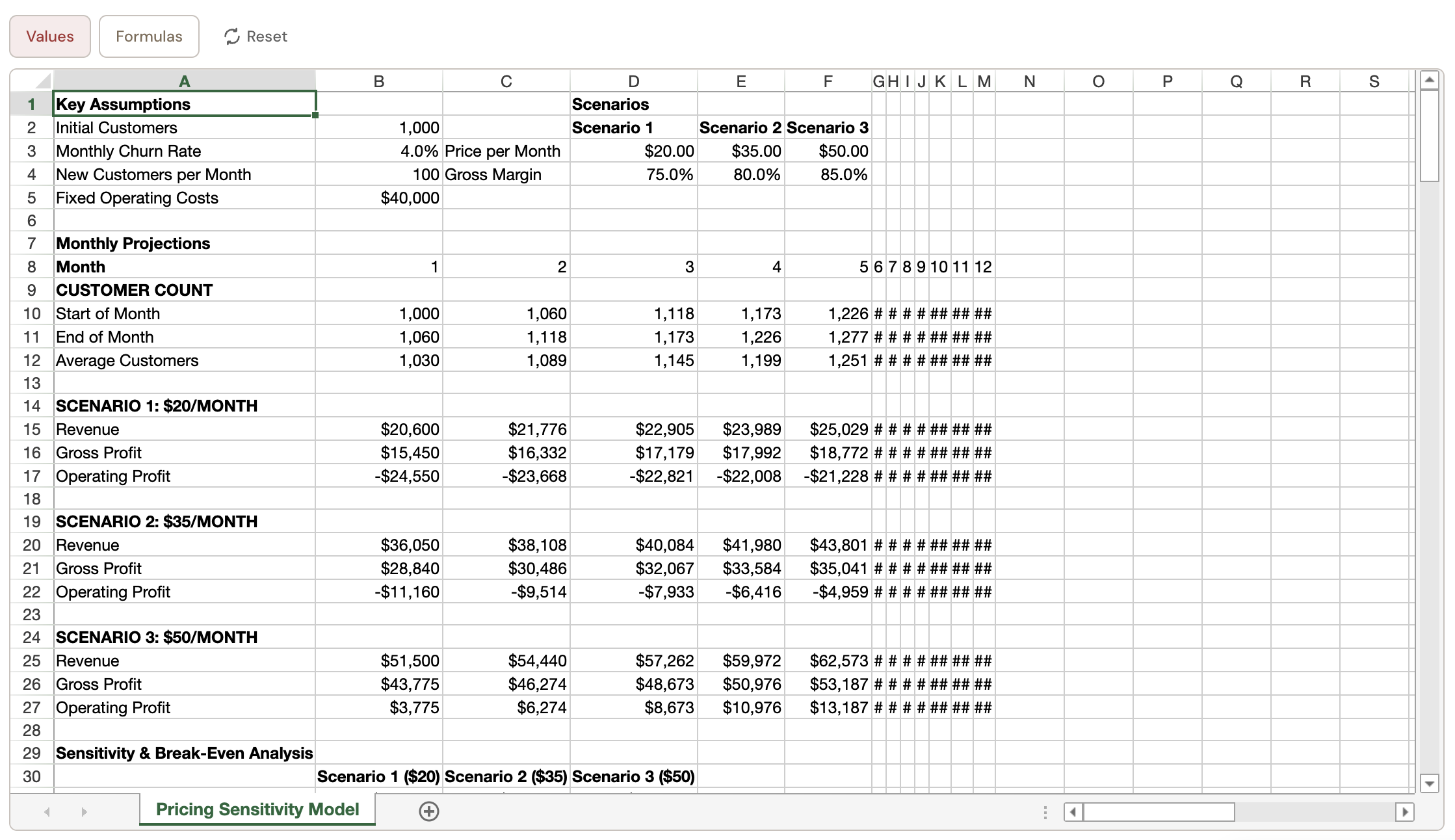}
    \caption{A model response to the ``Corporate Finance \& FP\&A'' prompt in Appx.~\ref{app:category_examples}.}
\end{figure}

\subsubsection{Creative \& Generative}

\begin{figure}[H]
    \centering
    \includegraphics[width=0.8\textwidth]{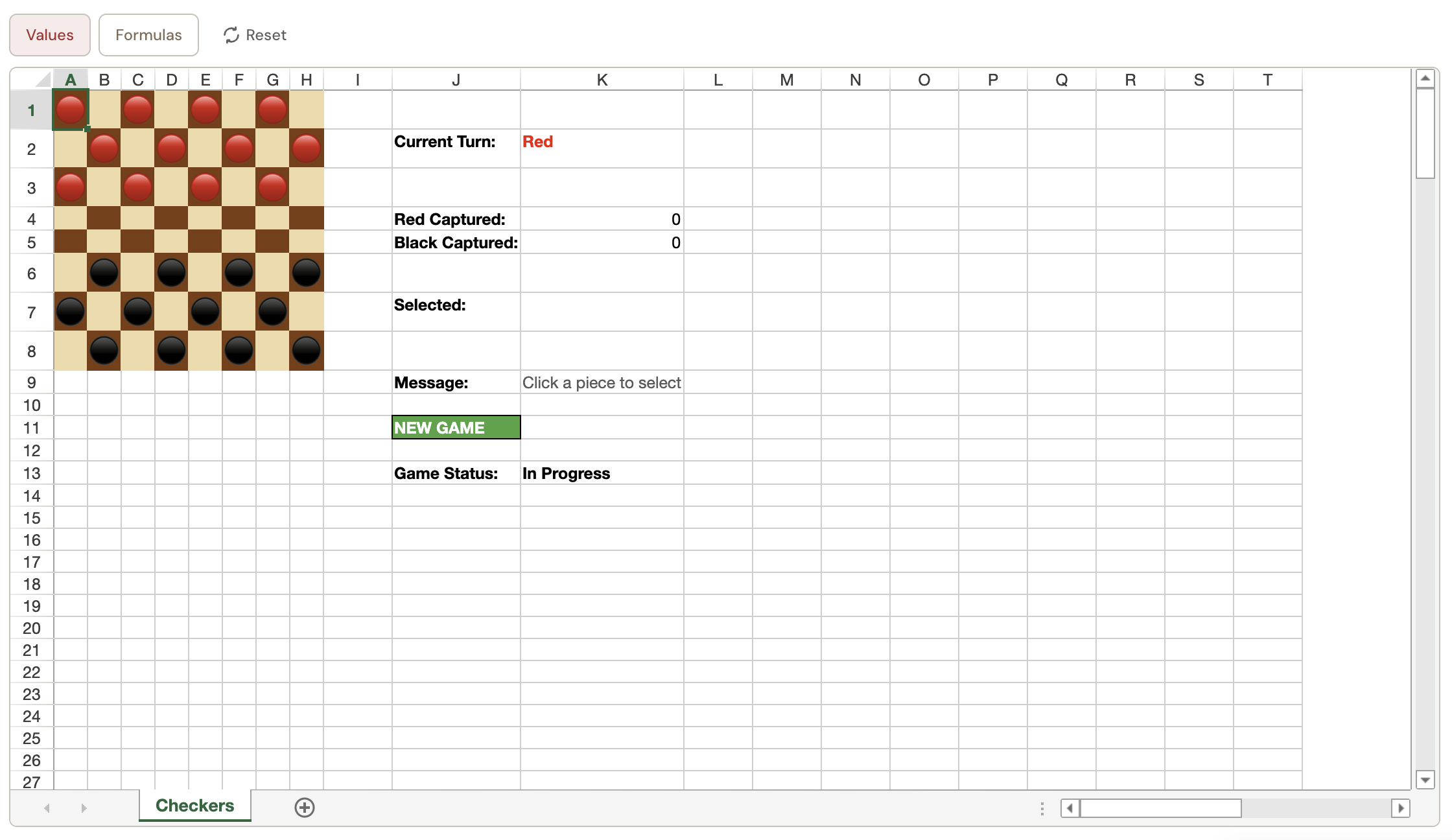}
    \caption{A model response to the ``Creative \& Generative'' prompt in Appx.~\ref{app:category_examples}.}
\end{figure}

\begin{figure}[H]
    \centering
    \includegraphics[width=0.8\textwidth]{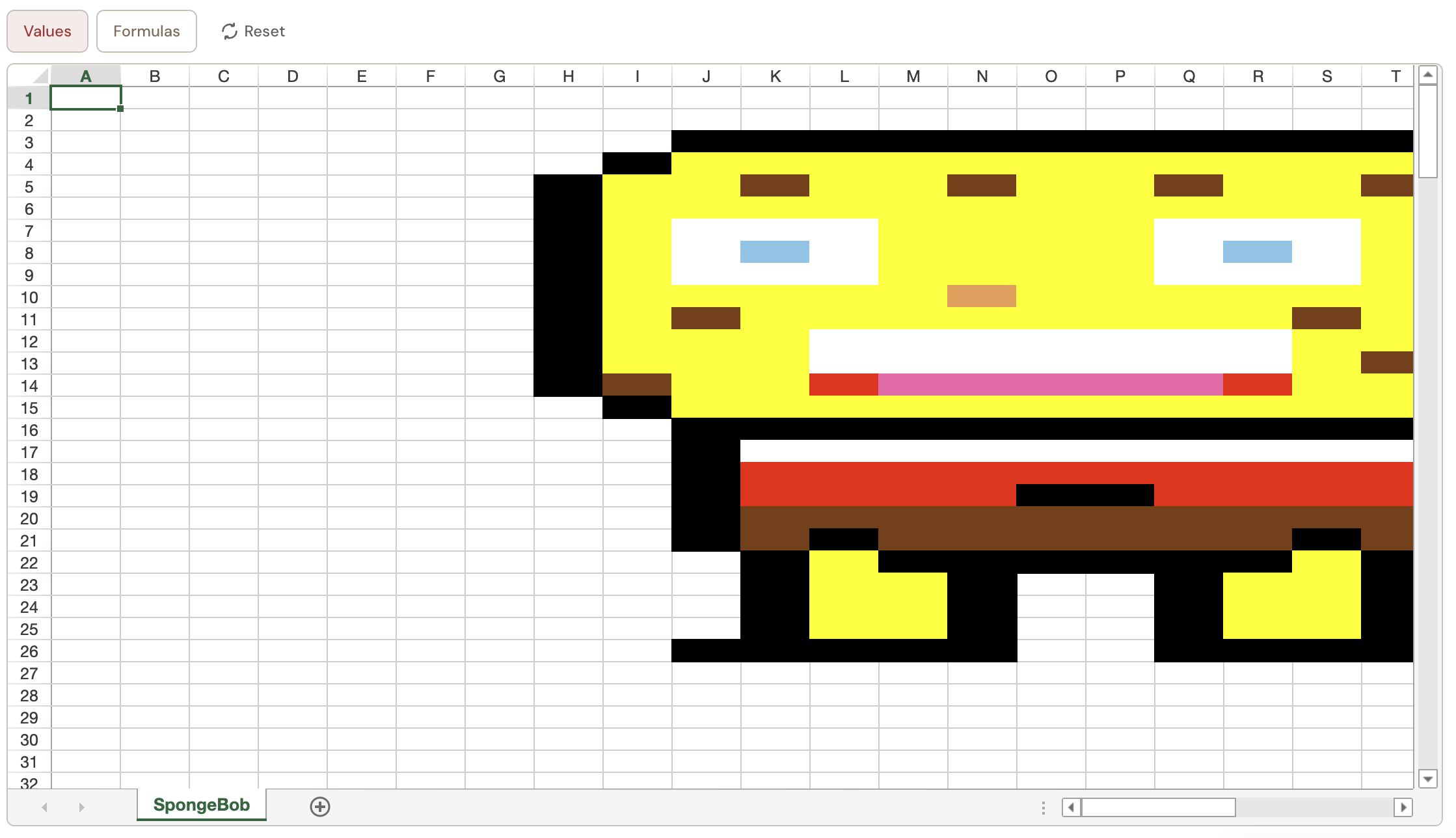}
    \caption{A model response to the prompt ``draw a spongebob fully colored and not exceeding a 50 x 50 size''.}
\end{figure}

\subsubsection{Operations \& Supply Chain}

\begin{figure}[H]
    \centering
    \includegraphics[width=0.8\textwidth]{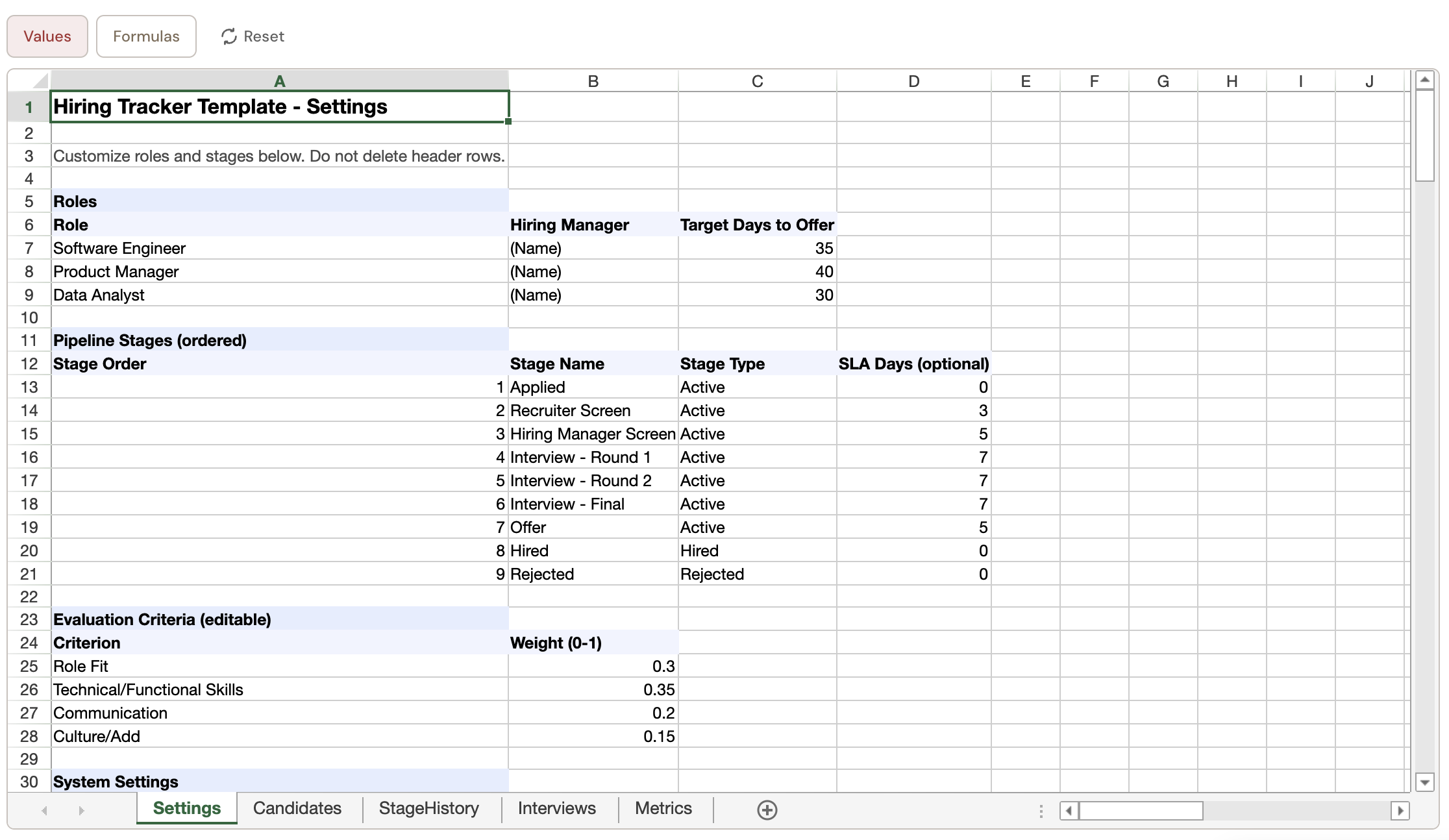}
    \caption{A model response to the ``Operations \& Supply Chain'' prompt in Appx.~\ref{app:category_examples}.}
\end{figure}

\subsubsection{Professional Finance}
\begin{figure}[H]
    \centering
    \includegraphics[width=0.8\textwidth]{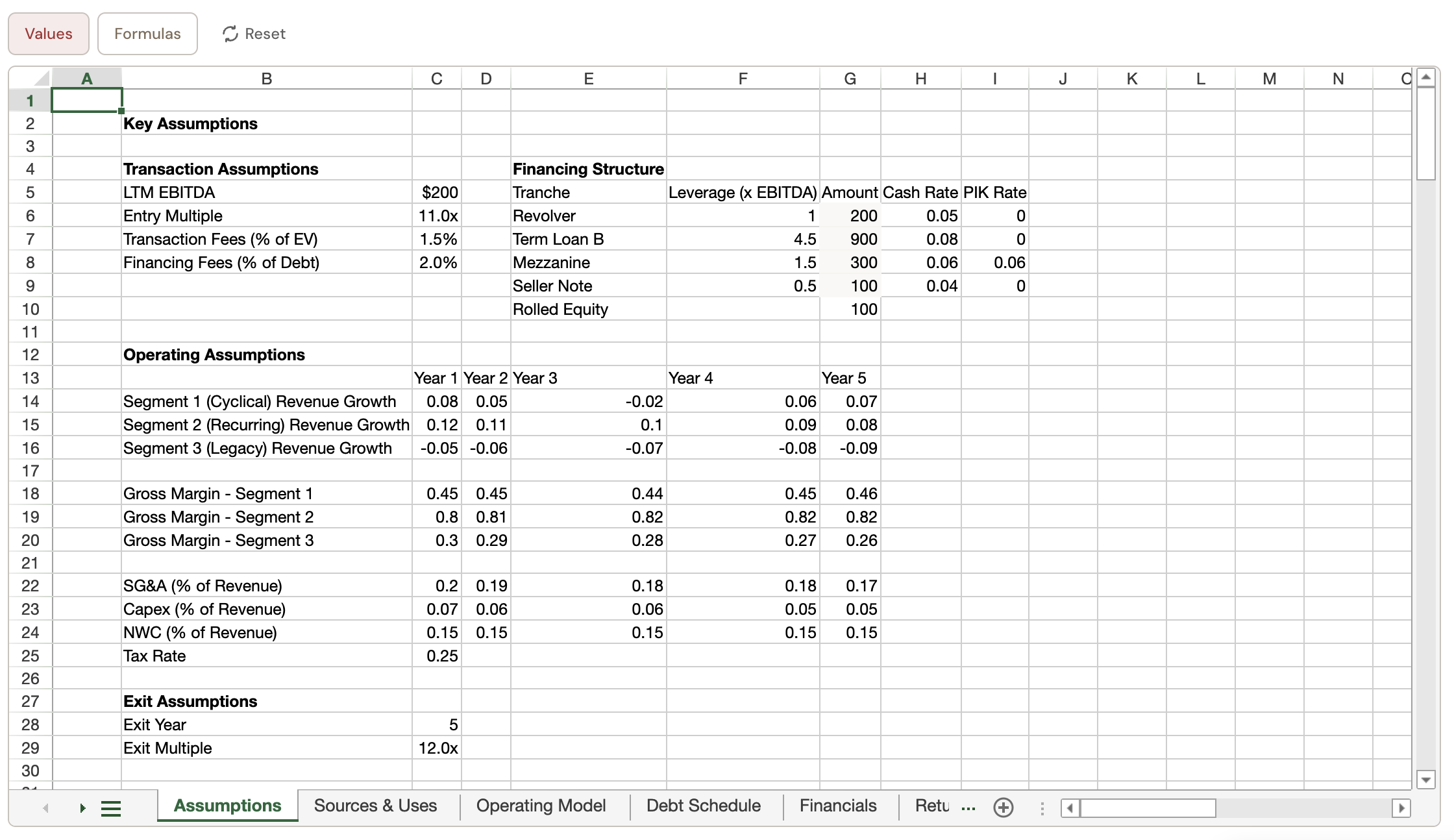}
    \caption{A model response to the ``Professional Finance'' prompt in Appx.~\ref{app:category_examples}.}
\end{figure}

\subsubsection{SMB \& Personal}
\begin{figure}[H]
    \centering
    \includegraphics[width=0.8\textwidth]{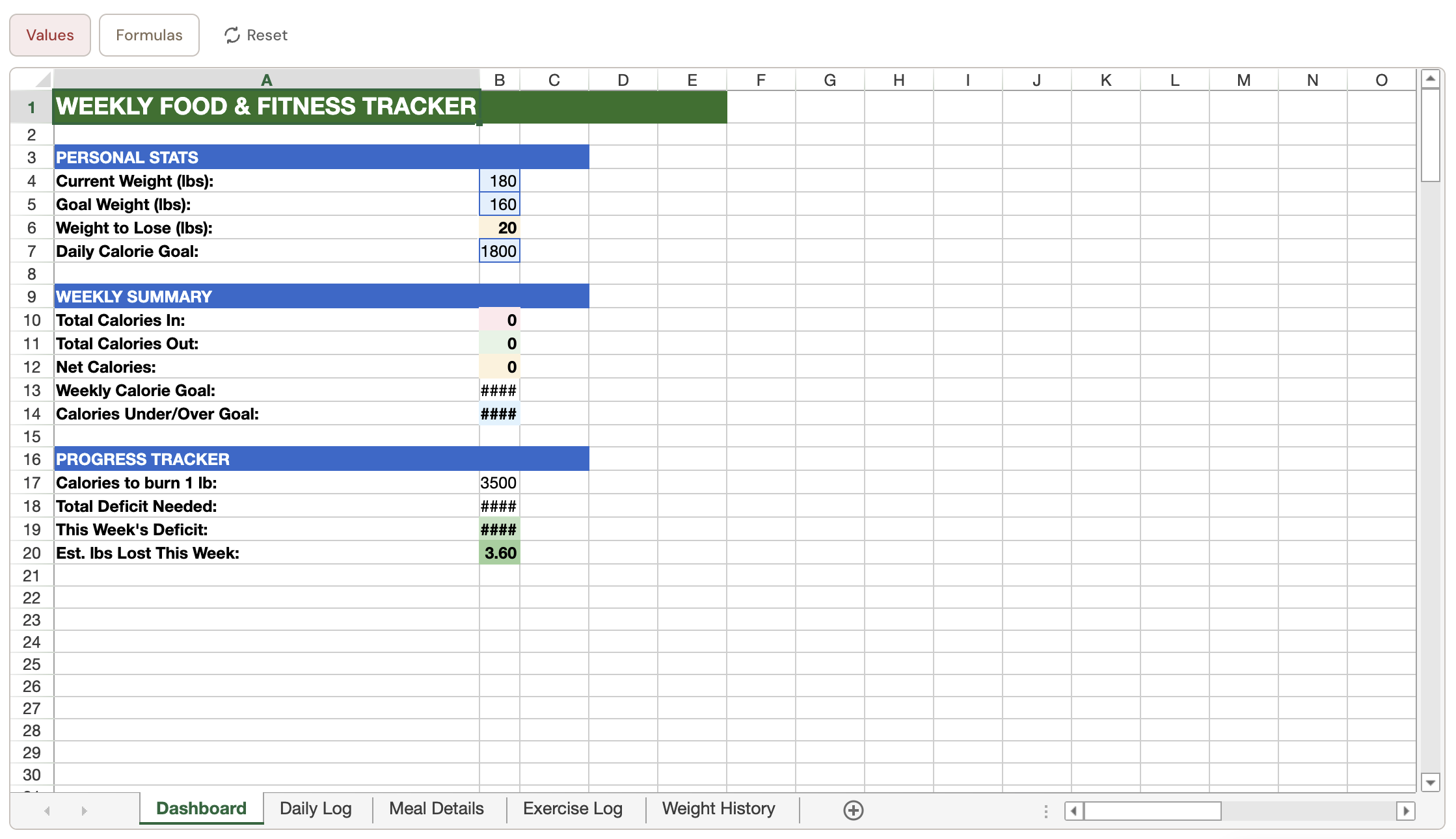}
    \caption{A model response to the ``SMB \& Personal'' prompt in Appx.~\ref{app:category_examples}.}
\end{figure}

\section{Expanded Model Coefficients Results}
\label{appx:extra_results_ci_etc}

\begin{table}[ht]
\centering
\small
\caption{Model ratings with 95\% confidence intervals. Elo: bootstrap percentile intervals (asymmetric). Bradley-Terry: analytic intervals (symmetric, shown as $\pm$). Rows sorted by BT rating; the Elo column is therefore not monotonic. $^{*}$GPT-4o is the reference anchor, fixed at 1000.}
\label{tab:model_ratings_ci_expanded}
\begin{tabular}{lrrcc}
\toprule
Model & Matches & Decisive & Elo score & BT coefficient \\
\midrule
Claude Opus 4.5    & 582 & 542 & $1535.20^{+108.67}_{-95.10}$  & $1550.49 \pm 54.54$  \\
Claude Sonnet 4.5  & 968 & 873 & $1477.69^{+112.27}_{-111.34}$ & $1427.44 \pm 43.92$  \\
Claude Opus 4.1    & 875 & 802 & $1442.75^{+111.20}_{-109.19}$ & $1405.77 \pm 44.24$  \\
Gemini 3 Pro       & 607 & 557 & $1308.81^{+111.40}_{-115.60}$ & $1324.75 \pm 48.11$  \\
GPT-5.2            & 405 & 352 & $1260.77^{+120.78}_{-93.78}$  & $1296.88 \pm 54.99$  \\
Gemini 2.5 Pro     & 916 & 813 & $1208.66^{+110.46}_{-106.80}$ & $1280.31 \pm 42.51$  \\
Grok 4.1 Fast      & 322 & 267 & $1242.64^{+116.24}_{-92.31}$  & $1255.69 \pm 59.17$  \\
Gemini 2.5 Flash   & 415 & 388 & $1146.95^{+116.52}_{-111.58}$ & $1255.64 \pm 49.30$  \\
GPT-5             & 761 & 667 & $1148.76^{+97.34}_{-99.16}$   & $1186.28 \pm 42.87$  \\
GPT-5.1           & 586 & 484 & $1178.29^{+104.46}_{-109.78}$ & $1159.81 \pm 49.23$  \\
Grok 4            & 384 & 350 & $1003.19^{+120.31}_{-120.57}$ & $1145.43 \pm 50.89$  \\
Grok Code Fast 1   & 556 & 463 & $1160.10^{+108.53}_{-99.22}$  & $1089.54 \pm 51.01$  \\
GPT-4o            & 789 & 675 & $1000.00$\textsuperscript{*}  & $1000.00$\textsuperscript{*} \\
Kimi K2 Instruct   & 276 & 208 & $1056.89^{+105.58}_{-113.20}$ & $977.15 \pm 74.08$   \\
Qwen3 30B          & 125 &  83 & $963.30^{+106.06}_{-86.77}$   & $691.80 \pm 157.04$  \\
Llama 4 Maverick   & 134 &  89 & $898.34^{+85.73}_{-88.78}$    & $632.63 \pm 167.40$  \\
\bottomrule
\end{tabular}
\end{table}

\section{Spreadsheet Features Appendix}
\label{app:spreadsheet_features}

Table~\ref{tab:features} contains descriptions of all 29 spreadsheet features used as covariates in the Bradley-Terry model. Features are sorted into four categories spanning formula quality, content, formatting, and structure.

Table~\ref{tab:feature-effects-all} contains feature effects on win probability for all prompts, for all 29 features.

\begin{table}[th!]
\centering
\caption{Spreadsheet features used as covariates in the Bradley-Terry model, grouped by category.}
\label{tab:features}
\small
\begin{tabular}{lll}
\toprule
\textbf{Category} & \textbf{Feature} & \textbf{Description} \\
\midrule
\multirow{6}{*}{\textbf{Formula Quality}}
& \texttt{compute\_error\_rate} & Formula error rate \\
& \texttt{compute\_pct\_numeric} & Numeric cell ratio \\
& \texttt{log\_distinct\_functions} & Function variety \\
& \texttt{log\_num\_lookups} & Lookup function count \\
& \texttt{log\_num\_conditionals} & Conditional function count \\
& \texttt{pct\_formulas\_with\_literals} & Embedded constants \\
\midrule
\multirow{3}{*}{\textbf{Content}}
& \texttt{pct\_text} & Text cell ratio \\
& \texttt{pct\_formula} & Formula cell ratio \\
& \texttt{log\_total\_text\_tokens} & Text word count \\
\midrule
\multirow{9}{*}{\textbf{Formatting}}
& \texttt{pct\_fill} & Background fill ratio \\
& \texttt{pct\_bold} & Bold text ratio \\
& \texttt{has\_border} & Border presence \\
& \texttt{pct\_number\_format} & Number formatting ratio \\
& \texttt{distinct\_font\_sizes} & Font size variety \\
& \texttt{pct\_font\_color} & Font color ratio \\
& \texttt{log\_distinct\_font\_colors} & Font color variety \\
& \texttt{distinct\_fills} & Fill color variety \\
& \texttt{finance\_color\_convention} & Color convention score \\
\midrule
\multirow{11}{*}{\textbf{Structure}}
& \texttt{log\_row\_count} & Row count \\
& \texttt{log\_col\_count} & Column count \\
& \texttt{log\_aspect\_ratio} & Sheet aspect ratio \\
& \texttt{cell\_density} & Non-empty cell ratio \\
& \texttt{log\_num\_blank\_rows} & Blank row count \\
& \texttt{num\_single\_cell\_rows} & Single-cell rows \\
& \texttt{num\_tables} & Table count \\
& \texttt{has\_parallel\_tables} & Side-by-side tables \\
& \texttt{avg\_tables\_per\_sheet} & Tables per sheet \\
& \texttt{largest\_table\_pct} & Largest table share \\
& \texttt{log\_table\_size\_variance} & Table size variance \\
\bottomrule
\end{tabular}
\end{table}

\begin{table}[htbp]
\centering
\small
\caption{Feature Effects on Win Probability (All Prompts). Asterisks denote statistical significance (*$p<0.05$ (no multiple tests correction), **significant under Benjamini-Hochberg adjustment, ***significant under Benjamini-Yekutieli adjustment}
\label{tab:feature-effects-all}
\begin{tabular}{@{}lrr@{}}
\toprule
\textbf{Feature} & \textbf{Coef.} & \textbf{$p$-value} \\
\midrule
\texttt{pct\_text} & $+1.562$ & $<0.001$*** \\
\texttt{compute\_error\_rate} & $-1.338$ & $<0.001$*** \\
\texttt{pct\_fill} & $+1.150$ & $<0.001$*** \\
\texttt{compute\_pct\_numeric} & $+1.020$ & 0.002*** \\
\texttt{log\_aspect\_ratio} & $-0.814$ & 0.010** \\
\texttt{pct\_formula} & $+0.711$ & 0.096 \\
\texttt{log\_col\_count} & $+0.725$ & $<0.001$*** \\
\texttt{pct\_number\_format} & $+0.657$ & $<0.001$*** \\
\texttt{pct\_font\_color} & $+0.592$ & 0.152 \\
\texttt{finance\_color\_conv.} & $+0.558$ & 0.094 \\
\texttt{largest\_table\_pct} & $-0.563$ & 0.013** \\
\texttt{has\_border} & $+0.312$ & 0.005*** \\
\texttt{cell\_density} & $+0.303$ & 0.200 \\
\texttt{log\_row\_count} & $+0.249$ & 0.096 \\
\texttt{log\_num\_blank\_rows} & $-0.248$ & 0.002*** \\
\texttt{has\_parallel\_tables} & $-0.214$ & 0.026** \\
\texttt{log\_distinct\_functions} & $-0.211$ & 0.026** \\
\texttt{log\_total\_text\_tokens} & $+0.167$ & 0.014** \\
\texttt{log\_distinct\_font\_colors} & $+0.153$ & 0.148 \\
\texttt{pct\_formulas\_w\_literals} & $+0.114$ & 0.547 \\
\texttt{avg\_tables\_per\_sheet} & $+0.104$ & $<0.001$*** \\
\texttt{distinct\_font\_sizes} & $+0.087$ & 0.078 \\
\texttt{log\_table\_size\_variance} & $+0.050$ & 0.003*** \\
\texttt{log\_num\_conditionals} & $+0.037$ & 0.283 \\
\texttt{num\_single\_cell\_rows} & $-0.027$ & 0.005*** \\
\texttt{log\_num\_lookups} & $+0.017$ & 0.730 \\
\texttt{num\_tables} & $-0.013$ & 0.058 \\
\texttt{distinct\_fills} & $+0.013$ & 0.252 \\
\texttt{pct\_bold} & $-0.040$ & 0.880 \\
\bottomrule
\end{tabular}
\end{table}

\begin{figure}[ht!]
    \centering
    \includegraphics[width=\columnwidth]{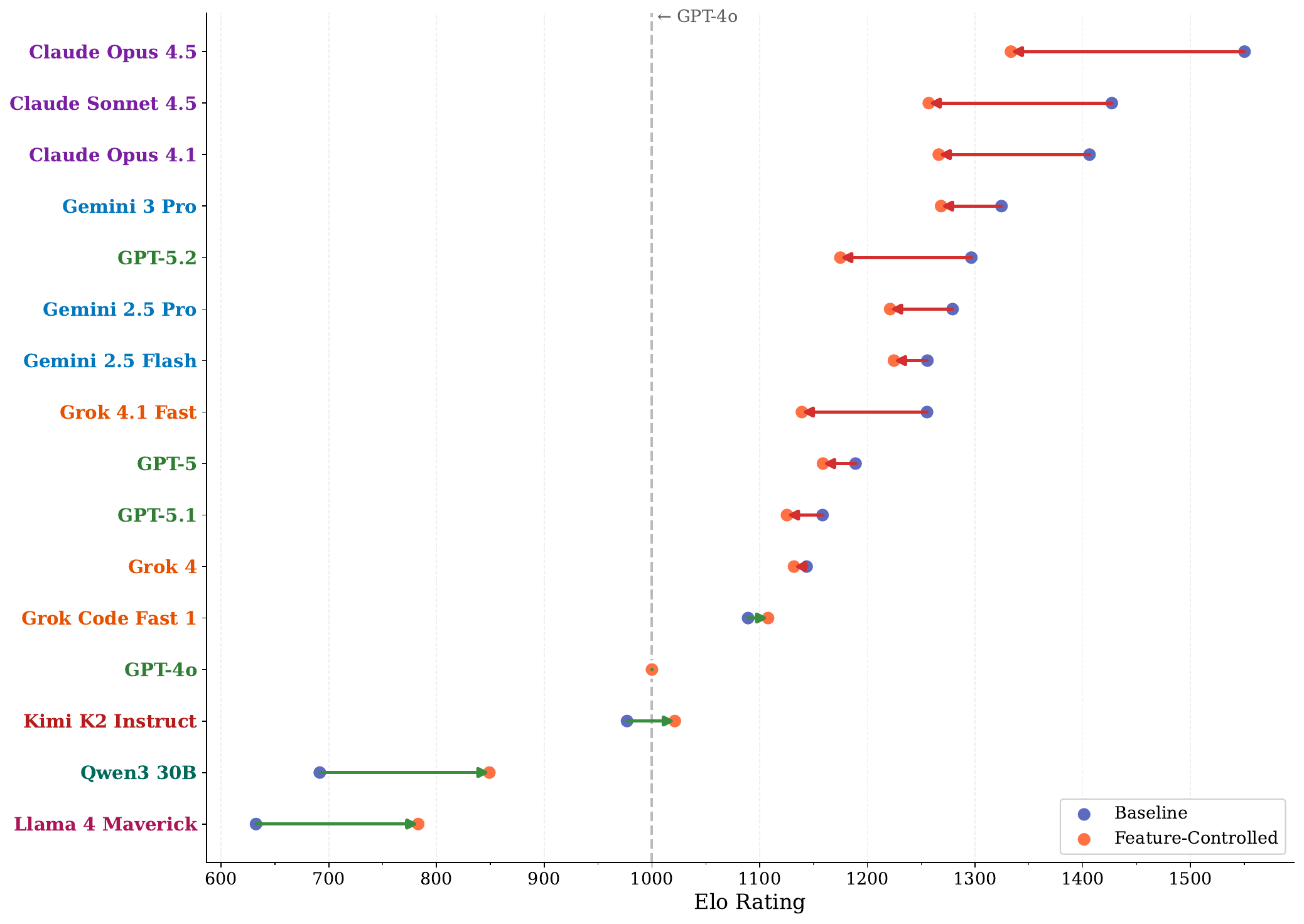}
    \caption{Elo ratings trend inwards after feature adjustment.}
    \label{fig:effect_style_controls}
\end{figure}

\Cref{fig:effect_style_controls} shows the effects of controlling for all features on Elo ratings.

\section{Feature Coefficients By Category}
\label{app:feature_coefs_category_expanded}

\cref{tab:feature_coefficients_category} contains feature coefficients and p-values for our set of 29 features, across our six prompt categories. We use a single finance category for this analysis, merging professional finance and corporate and FP\&A categories.
\begin{table*}[ht]
\caption{Feature coefficients and p-values across prompt categories. Bold indicates raw p-values $< 0.05$, though we note that many of these features do not survive Benjamini-Hochberg correction, unsurprising due to the relatively small sample sizes. Coefficients represent the effect on the log-odds of winning.}
\label{tab:feature_coefficients_category}
\centering
\footnotesize
\setlength{\tabcolsep}{4pt}
\begin{tabular}{l|rr|rr|rr|rr|rr}
\toprule
& \multicolumn{2}{c|}{\textbf{Creative \&}} & \multicolumn{2}{c|}{\textbf{Finance}} & \multicolumn{2}{c|}{\textbf{Academic \&}} & \multicolumn{2}{c|}{\textbf{SMB \&}} & \multicolumn{2}{c}{\textbf{Operations \&}} \\
& \multicolumn{2}{c|}{\textbf{Generative}} & \multicolumn{2}{c|}{\textbf{(Prof. + Corp.)}} & \multicolumn{2}{c|}{\textbf{Research}} & \multicolumn{2}{c|}{\textbf{Personal}} & \multicolumn{2}{c}{\textbf{Supply Chain}} \\
\addlinespace[2pt]
\textbf{Feature} & $\beta$ & $p$ & $\beta$ & $p$ & $\beta$ & $p$ & $\beta$ & $p$ & $\beta$ & $p$ \\
\midrule
\multicolumn{11}{l}{\textbf{Formula Quality}} \\
\texttt{compute\_error\_rate} & $-0.90$ & .273 & $-1.36$ & .231 & $-1.01$ & .660 & $+0.66$ & .594 & $\mathbf{-2.34}$ & $\mathbf{.042}$ \\
\texttt{compute\_pct\_numeric} & $+1.02$ & .169 & $+1.44$ & .141 & $+3.65$ & .071 & $\mathbf{+3.16}$ & $\mathbf{.002}$ & $+0.18$ & .863 \\
\texttt{log\_distinct\_functions} & $+0.22$ & .601 & $-0.20$ & .264 & $-0.44$ & .256 & $-0.14$ & .585 & $-0.58$ & .061 \\
\texttt{log\_num\_lookups} & $\mathbf{-0.45}$ & $\mathbf{.004}$ & $+0.07$ & .464 & $-0.22$ & .215 & $+0.05$ & .704 & $+0.03$ & .849 \\
\texttt{log\_num\_conditionals} & $-0.02$ & .909 & $+0.02$ & .735 & $-0.11$ & .489 & $+0.04$ & .683 & $\mathbf{+0.29}$ & $\mathbf{.010}$ \\
\texttt{pct\_formulas\_with\_literals} & $+0.29$ & .568 & $-0.19$ & .688 & $+0.04$ & .972 & $-0.61$ & .158 & $+0.78$ & .165 \\
\midrule
\multicolumn{11}{l}{\textbf{Content}} \\
\texttt{pct\_text} & $\mathbf{+2.35}$ & $\mathbf{.027}$ & $+0.41$ & .774 & $+1.38$ & .593 & $\mathbf{+3.52}$ & $\mathbf{.007}$ & $\mathbf{+3.41}$ & $\mathbf{.025}$ \\
\texttt{pct\_formula} & $+0.96$ & .404 & $-0.35$ & .730 & $+2.02$ & .328 & $\mathbf{+2.68}$ & $\mathbf{.036}$ & $+1.46$ & .258 \\
\texttt{log\_total\_text\_tokens} & $+0.15$ & .212 & $+0.33$ & .149 & $+0.53$ & .256 & $+0.10$ & .663 & $+0.08$ & .824 \\
\midrule
\multicolumn{11}{l}{\textbf{Formatting}} \\
\texttt{pct\_fill} & $\mathbf{+1.17}$ & $\mathbf{.020}$ & $+0.65$ & .530 & $\mathbf{+3.17}$ & $\mathbf{.040}$ & $+1.45$ & .059 & $-0.57$ & .451 \\
\texttt{pct\_bold} & $-0.37$ & .478 & $+0.60$ & .397 & $+0.37$ & .858 & $\mathbf{-1.97}$ & $\mathbf{.008}$ & $-2.02$ & .051 \\
\texttt{has\_border} & $\mathbf{-0.87}$ & $\mathbf{.023}$ & $\mathbf{+0.57}$ & $\mathbf{.013}$ & $-0.37$ & .569 & $+0.10$ & .698 & $\mathbf{+0.71}$ & $\mathbf{.028}$ \\
\texttt{pct\_number\_format} & $-1.13$ & .430 & $\mathbf{+0.61}$ & $\mathbf{.046}$ & $\mathbf{-5.38}$ & $\mathbf{.041}$ & $+0.99$ & .065 & $+1.63$ & .164 \\
\texttt{distinct\_font\_sizes} & $+0.16$ & .317 & $-0.01$ & .904 & $+0.02$ & .946 & $+0.14$ & .275 & $+0.03$ & .883 \\
\texttt{pct\_font\_color} & $+1.25$ & .119 & $+0.23$ & .898 & $+0.88$ & .819 & $+0.17$ & .902 & $-0.42$ & .762 \\
\texttt{log\_distinct\_font\_colors} & $+0.15$ & .660 & $+0.30$ & .223 & $+0.53$ & .493 & $+0.06$ & .805 & $-0.08$ & .796 \\
\texttt{distinct\_fills} & $+0.01$ & .670 & $-0.04$ & .299 & $+0.25$ & .055 & $+0.07$ & .217 & $-0.06$ & .238 \\
\texttt{finance\_color\_convention} & $+0.74$ & .390 & $\mathbf{+1.63}$ & $\mathbf{.022}$ & $-0.24$ & .889 & $-1.45$ & .152 & $-0.23$ & .836 \\
\midrule
\multicolumn{11}{l}{\textbf{Structure}} \\
\texttt{log\_row\_count} & $\mathbf{+2.13}$ & $\mathbf{.044}$ & $+0.41$ & .370 & $-0.71$ & .277 & $+0.94$ & .057 & $+0.12$ & .824 \\
\texttt{log\_col\_count} & $+0.05$ & .958 & $+0.15$ & .735 & $+1.19$ & .259 & $+0.73$ & .174 & $+1.14$ & .118 \\
\texttt{log\_aspect\_ratio} & $+0.20$ & .902 & $-0.49$ & .624 & $-1.62$ & .361 & $+0.40$ & .664 & $+0.19$ & .857 \\
\texttt{cell\_density} & $\mathbf{+2.22}$ & $\mathbf{.012}$ & $-0.13$ & .811 & $+1.80$ & .083 & $-0.79$ & .186 & $+0.62$ & .464 \\
\texttt{log\_num\_blank\_rows} & $-0.65$ & .108 & $-0.17$ & .316 & $+0.41$ & .358 & $-0.31$ & .160 & $-0.45$ & .091 \\
\texttt{num\_single\_cell\_rows} & $-0.07$ & .153 & $-0.02$ & .402 & $+0.02$ & .776 & $-0.03$ & .160 & $+0.07$ & .335 \\
\texttt{num\_tables} & $-0.09$ & .558 & $-0.00$ & .996 & $+0.02$ & .762 & $\mathbf{-0.12}$ & $\mathbf{.002}$ & $-0.07$ & .117 \\
\texttt{has\_parallel\_tables} & $+0.41$ & .335 & $-0.20$ & .308 & $-0.02$ & .955 & $\mathbf{-0.57}$ & $\mathbf{.016}$ & $-0.26$ & .411 \\
\texttt{avg\_tables\_per\_sheet} & $\mathbf{+0.50}$ & $\mathbf{.008}$ & $+0.03$ & .617 & $+0.06$ & .718 & $\mathbf{+0.26}$ & $\mathbf{.005}$ & $\mathbf{+0.34}$ & $\mathbf{.044}$ \\
\texttt{largest\_table\_pct} & $+0.92$ & .369 & $\mathbf{-1.00}$ & $\mathbf{.030}$ & $-0.76$ & .521 & $+0.42$ & .468 & $\mathbf{-2.17}$ & $\mathbf{.015}$ \\
\texttt{log\_table\_size\_variance} & $+0.00$ & .937 & $+0.07$ & .078 & $+0.05$ & .426 & $+0.02$ & .646 & $-0.03$ & .431 \\
\bottomrule
\end{tabular}
\end{table*}

\section{Category Model Rankings Changes}
\label{appx:rank-changes}

In Academic \& Research prompts, we see the most dramatic ranking perturbation in our study (see Tab.~\ref{tab:academic-rankings} in Appx.~\ref{appx:academic-rank-changes}). Claude Opus~4.5 drops from 1st to 9th place ($-236$ Elo), while Grok~4, which already had an unusually high baseline, ascends to the top ($+149$ Elo) and GPT-5.1 gains 228 points. Only two features achieve significance in this domain, reported in Tab~\ref{tab:feature_coefficients_category}, but the large negative coefficient, $-5.38$ ($p = 0.04$), for \texttt{pct\_number\_format} is noteworthy -- Claude's heavy use of formatting negatively affects perceived negatively in this domain.

In contrast, in the Finance domain (Tab.~\ref{tab:finance-rankings} in Appx.~\ref{appx:finance-rank-changes}), four features achieve significance (Tab.~\ref{tab:feature_coefficients_category}), three of which reflect professional financial modeling conventions. The strongest predictor is \texttt{finance\_color\_convention\_score}, which is not statistically significant arena-wide ($p = 0.09$) but has a coefficient of $+1.63$ ($p = 0.02$) for the Finance domain. We note that, though alignment with color conventions is simple to check for programmatically, full evaluation of adherence to financial modeling conventions is more challenging; see \S\ref{subsec:finance_study} for an expert evaluation study. Tab.~\ref{tab:finance-rankings} contains ranking changes for models over both finance categories. 

\section{Finance Category Model Rankings Change}
\label{appx:finance-rank-changes}

\cref{tab:finance-rankings} contains ranking changes for models over both finance categories. 

\begin{table}[htbp!]
\caption{Model Rankings After Feature Adjustments: Professional Finance \& Corporate FP\&A}
\label{tab:finance-rankings}
\centering
\small
\begin{tabular}{@{}lrrrr@{}}
\toprule
\textbf{Model} & \textbf{Elo} & \textbf{Ctrl Elo} & \textbf{$\Delta$Elo} & \textbf{$\Delta$Rank} \\
\midrule
Claude Opus 4.5 & 1678 & 1395 & $-283$ & $0$ \\
Claude Opus 4.1 & 1586 & 1376 & $-209$ & $0$ \\
Claude Sonnet 4.5 & 1580 & 1334 & $-247$ & $0$ \\
Gemini 3 Pro & 1502 & 1312 & $-190$ & $0$ \\
Gemini 2.5 Flash & 1448 & 1294 & $-154$ & $+2$ \\
GPT-5.2 & 1493 & 1293 & $-200$ & $-1$ \\
Gemini 2.5 Pro & 1453 & 1256 & $-198$ & $-1$ \\
GPT-5 & 1318 & 1229 & $-89$ & $+1$ \\
GPT-5.1 & 1293 & 1172 & $-121$ & $+1$ \\
Grok Code Fast 1 & 1208 & 1157 & $-51$ & $+1$ \\
Grok 4.1 Fast & 1392 & 1152 & $-240$ & $-3$ \\
Kimi K2 Instruct & 1089 & 1088 & $-1$ & $0$ \\
GPT-4o & 1000 & 1000 & $0$ & $0$ \\
\bottomrule
\end{tabular}
\end{table}

\section{Academic \& Research Category Model Rankings Change}
\label{appx:academic-rank-changes}

\cref{tab:finance-rankings} contains ranking changes for models over both finance categories. 

\begin{table}[htbp]
\caption{Model Rankings After Feature Adjustment: Academic \& Research}
\label{tab:academic-rankings}
\centering
\small
\begin{tabular}{@{}lrrrr@{}}
\toprule
\textbf{Model} & \textbf{Elo} & \textbf{Ctrl Elo} & \textbf{$\Delta$Elo} & \textbf{$\Delta$Rank} \\
\midrule
Grok 4 & 1481 & 1630 & $+149$ & $+1$ \\
GPT-5.1 & 1298 & 1526 & $+228$ & $+4$ \\
Gemini 3 Pro & 1305 & 1457 & $+152$ & $+2$ \\
GPT-5 & 1257 & 1449 & $+192$ & $+5$ \\
Gemini 2.5 Flash & 1297 & 1432 & $+135$ & $+2$ \\
Claude Opus 4.1 & 1429 & 1414 & $-15$ & $-2$ \\
Gemini 2.5 Pro & 1283 & 1367 & $+84$ & $+1$ \\
Claude Sonnet 4.5 & 1446 & 1360 & $-85$ & $-5$ \\
Claude Opus 4.5 & 1527 & 1291 & $-236$ & $-8$ \\
Grok Code Fast 1 & 1141 & 1246 & $+105$ & $0$ \\
GPT-4o & 1000 & 1000 & $0$ & $0$ \\
\bottomrule
\end{tabular}
\end{table}

\newpage
\section{Failure Modes Analysis}
\label{appx:failure-modes}

See Table~\ref{tab:failure_rates} for full results from the study in \S\ref{subsec:loss_buckets}.

\subsection{Methodological Details}
\paragraph{Category Discovery. } We follow BERTopic \citep{grootendorst2022bertopicneuraltopicmodeling} to design a data-driven discovery pipeline to surface natural failure patterns from the arena corpus. We first generate open-ended failure rationales for a sample of 260 decisive battles (stratified across prompt category, losing model, and prompt complexity). For each battle, the \texttt{gpt-5-mini} 
judge receives JSON representations of both candidate spreadsheets along with the prompt text and winner designation, and produces a structured assessment of the losing spreadsheet's shortcomings.

We then embed these rationales using OpenAI's \texttt{text-embedding-3-small} model, reduce dimensionality with UMAP (5 components), and cluster via HDBSCAN with a minimum cluster size of 10. Central rationales from each cluster are fed to GPT-5 to generate descriptive category names and definitions. This pipeline yields 9 natural clusters, which we use as a starting point for the final hand-curated taxonomy of 7 buckets. 

\paragraph{Judging Method.}                         
After establishing our taxonomy, we apply our \texttt{gpt-5-mini} judge to each decisive arena battle, where one output was preferred over the other. The \texttt{gpt-5-mini} judge receives the original prompt and both full candidates as input. A system prompt (see Appendix~\ref{appx:judge_si}) provides all 8 category definitions with examples and instructs the judge to tag the losing spreadsheet with all relevant error categories, requiring clear evidence for each tag. The judge returns a structured JSON object containing the list of applicable category IDs and a 2-3 sentence rationale citing specific evidence, with example rationales in Appendix~\ref{app:loss-bucket-judge-samples}. This multi-label design captures failure co-occurrence.

\subsection{Sample Loss Categorization Judge Rationales}
\label{app:loss-bucket-judge-samples}

Table~\ref{tab:bucket-samples} contains sample LLM judge rationales for bucket categorizations.

\begin{longtable}{p{3.5cm} p{11cm}}
\caption{Sample LLM judge rationales for bucket categorizations.}
\label{tab:bucket-samples}
\\
\textbf{Loss Bucket} & \textbf{Judge Rationale} \\
\midrule
\endfirsthead
\toprule
\textbf{Loss Bucket} & \textbf{Judge Rationale} \\
\midrule
\endhead
\midrule
\multicolumn{2}{r}{\textit{Continued on next page}} \\
\endfoot
\bottomrule
\endlastfoot
Non-functional &
Calculations contain pervasive formula errors caused by incorrect sheet references (e.g., \texttt{Calculations!B6..G6} and \texttt{B7..G7} use \texttt{'Assumptions.B6'} instead of \texttt{'Assumptions!B6'}), leaving key outputs non-functional. \\
\midrule
Spec Non-compliance &
The model fails the prompt requirement: the sensitivity table (\texttt{DCF!B43:F45}) produces enterprise-value outputs and is not converted to equity value per share (prompt requested equity value sensitivity). \\
\midrule
Integrity Failure &
Input assumptions are not single-sourced or consistently linked (hardcoded step-up and amortization values are placed as year values rather than centralized blue input cells). \\
\midrule
Numerical Computation Failure &
There is incorrect math in the implied share price: \texttt{Bridge!B11} and \texttt{Bridge!B17} multiply price by 10 (\texttt{B7/B9*10}), which is an obvious unit/signature error that produces wrong implied prices. \\
\midrule
Interpretability Failure &
Labels contradict layout (A1 = ``Quarter'' while rows are product lines), assumptions and calculations aren't separated, making the model hard to audit. \\
\midrule
Low User Value &
It provides little user value---no translations, counts, or selection rationale so it's largely a wall of characters (shallow, low decision value). \\
\midrule
Presentation Deficiency &
Date cells are entered as plain text with formatting (\texttt{Assumptions!B4:B6}, \texttt{B11}) instead of true date types, and some number/date formatting is inconsistent with the requested conventions (e.g., days/years precision and long-date display), which lowers professional polish and increases risk of hidden errors. \\
\end{longtable}

\subsection{Loss Categorization Judge System Prompt}
\label{appx:judge_si}

\begin{mdblock}
You are a senior spreadsheet professional analyzing why a spreadsheet lost a head-to-head arena battle. You have deep expertise in spreadsheet modeling, financial analysis, and the technical craft of building production-quality workbooks.

A human reviewer compared two spreadsheet outputs built for the same task and chose one as better. Your job is to think deeply and tag all failure modes that genuinely contributed to the loss.

### SheetSpec Format

The spreadsheets are in SheetSpec@2 JSON:
- Each workbook has sheets, each with an array of cells
- Cell types: `text` (string), `number` (numeric), `formula` (Excel A1)
- Cells may have `style`: `fill`, `fontWeight`, `fontSize`, `numberFormat`, `fontColor`, `border`
- Sheets may have `namedRanges` and `conditionalFormats`

### Failure Categories

**[0] Noise / Unjudgeable**
Definition: The spreadsheet can't be meaningfully judged against the prompt. Use this tag ONLY in extreme cases.
Indicators:
  - File is empty or contains unrelated content
  - Prompt is incoherent or sheet content doesn't correspond to the prompt
  - Generation is truncated in a way that makes evaluation impossible

**[1] Broken / Non-functional**
Definition: The spreadsheet is unusable - the equivalent of 'code that doesn't compile.'
Indicators:
  - Pervasive #DIV/0!, #REF!, #NAME?, or #VALUE! errors across key output areas
  - Circular references that clearly prevent meaningful outputs
  - Key results are blank or invalid due to broken references

**[2] Prompt Miss / Incomplete Build**
Definition: The spreadsheet doesn't include the core deliverables the prompt requires. It might calculate 'something,' but not what was asked.
Indicators:
  - Missing required sections, tabs, scenarios, or time horizons
  - Wrong dimensionality (e.g., annual vs. monthly when prompt specifies otherwise)
  - Coverage too narrow - only a small subset of what was requested
  - Key required outputs (e.g., MOIC table, sensitivity analysis) are absent

**[3] Integrity / Architecture Failure**
Definition: The spreadsheet is structurally untrustworthy even if it looks plausible - not just wrong, but misleading or non-integrated.
Indicators:
  - Hardcoded 'checks' - status says PASS because the checker is fake
  - Key drivers are not linked to outputs - model doesn't respond to input changes
  - Mis-referenced ranges, duplicated drivers, unintentional circularity
  - 'Single source of truth' is violated - brittle and non-auditable

**[4] Incorrect Logic / Math**
Definition: The formulas are linked and the structure is real, but the underlying logic or math is wrong.
Indicators:
  - Constraints violated in the output
  - Totals don't tie or reconcile
  - Off-by-one, double-counting, or sign-convention errors
  - Scenario outputs don't match inputs or stated assumptions

**[5] Unclear Structure / Interpretability Failure**
Definition: The spreadsheet is hard to follow, teach from, or hand off to a collaborator.
Indicators:
  - Assumptions, calculations, and outputs are not clearly separated
  - Labels and numbers are misaligned or ambiguous
  - Not auditable by someone who didn't build it

**[6] Low User Value / Shallow**
Definition: The sheet may be correct and readable, but doesn't provide meaningful decision value.
Indicators:
  - 'Wall of numbers' with no interpretive scaffolding
  - No sensitivity analysis, summaries, or 'so what' takeaways
  - Accurate but unactionable - letter of the prompt without serving user intent

**[7] Presentation / Convention Deficiency**
Definition: The spreadsheet loses on polish and professional conventions.
Indicators:
  - Messy or inconsistent formatting (number formats, alignment, spacing)
  - Nonstandard accounting presentation
  - Missing visual hierarchy (no section headers, no input/calc color coding)
  - Visually inferior in a head-to-head comparison

### Rules

- Tag ALL categories (1-7) that genuinely apply and significantly contributed to the loss.
- Only tag a category if you see clear evidence for it.
- Do NOT tag everything - be honest and specific about categories that truly contributed to the loss.
- Tag 0 (Noise) ONLY if the output truly cannot be evaluated - empty, truncated, or completely unrelated to the prompt. If there is any substantive content to judge, do NOT use tag 0.

### Input

You will receive:
- The original task prompt
- Spreadsheet A (SheetSpec@2 JSON)
- Spreadsheet B (SheetSpec@2 JSON)
- Any formula errors detected by the evaluation engine
- Which spreadsheet the human reviewer chose as better

Analyze the losing spreadsheet and explain why it lost.

### Output Format

Return a JSON object with exactly this schema (no other text):

{"tags": [1, 3], "rationale": "2-3 sentences citing specific evidence."}

- `tags`: array of integer category IDs (0-7), sorted ascending
- `rationale`: brief explanation with cell references or structural observations
\end{mdblock}




\section{Finance Expert Evaluation: Protocol, Rubric, and Detailed Results}
\label{app:finance_evaluation}

In this section, we provide the full study protocol, rater instructions and scoring anchors, and detailed results.

\paragraph{Motivation.} Professional finance spreadsheets generally adhere to established modeling conventions. Where our programmatic and data-driven taxonomies operate at scale across all domains, this study applies an expert-designed rubric to assess whether arena outputs meet the professional standards required by domain practitioners. Our investigations reveal that LLMs tend to produce spreadsheets with poor grounding in established industry conventions for financial modeling, necessitating substantial manual revision before use in professional workflows.

\subsection{Study Design and Protocol}
We selected 25 battles with strict preference outcomes (excluding \texttt{Tie} and \texttt{Both are bad}), yielding 25 win-loss pairs (50 spreadsheets total). Battles were restricted to finance-domain prompts using manual labeling of seed prompts and k-NN classification for unlabeled submissions (\S\ref{sec:methodology}). Prompts span canonical financial workflows, including DCFs, LBOs, and distribution waterfalls. Five evaluators with at least two years of Excel-based financial modeling experience (investment banking and private equity backgrounds) rated the spreadsheets.

Each evaluator scored the same set of 50 tasks, presented in an independently shuffled order. Each prompt appears twice---once for the battle winner and once for the loser---and evaluators were fully blinded to model identity, battle outcome, and winner/loser status. All five evaluators rated all 50 tasks (a fully-crossed, all-IRR design), yielding 250 total evaluations and enabling inter-rater reliability analysis across the complete set. Before beginning, evaluators read a research brief covering the rubric, the key questions to ask while reviewing, and pacing guidance. To limit reviewer fatigue and preserve scoring consistency, evaluators were asked to work in batches (e.g., batches of ten) rather than scoring all tasks in a single sitting; each review took approximately 5--8 minutes.

\subsection{Evaluator Instructions}
For each assigned task, evaluators completed the following steps:
\begin{enumerate}[itemsep=0pt, topsep=0pt]
    \item \textbf{Read the prompt.} Understand what the spreadsheet was supposed to accomplish.
    \item \textbf{Download and open the Excel file.} Review it as you would any financial model---check formulas, structure, formatting.
    \item \textbf{Rate on 6 criteria (1--5 scale).} Score each dimension using the detailed rubric below.
    \item \textbf{Answer the circulation question (yes/no).} Indicate whether you would share this spreadsheet with a client or colleague (\cref{subsec:circulation}).
    \item \textbf{Add notes (optional but helpful).} Brief explanations of scores help us understand the reasoning.
    \item \textbf{Review and submit.} The overall rating is calculated automatically from the six dimension scores.
\end{enumerate}

\subsection{Rating Scale}
All dimensions use a 5-point Likert scale with consistent anchors. The scale is described in \cref{tab:likert_scale}.

\begin{table}[h]
\caption{Likert scale description.} 
\label{tab:likert_scale}
\centering
\small
\begin{tabular}{cl}
\toprule
\textbf{Score} & \textbf{General Definition} \\
\midrule
1 & Poor: Significant issues; unacceptable in professional context \\
2 & Below Average: Notable problems requiring substantial work to fix \\
3 & Acceptable: Meets minimum requirements; functional but not polished \\
4 & Good: Above average with only minor issues; professional quality \\
5 & Excellent: Exceptional quality exemplifying best practices \\
\bottomrule
\end{tabular}
\end{table}

\subsection{Evaluation Dimensions and Scoring Anchors}

\begin{table}[h]
\caption{\textbf{Dimension 1: Errors \& Accuracy.} \textit{Focus}: Formula correctness and absence of Excel errors. This criterion evaluates whether the spreadsheet is free from formula errors, Excel error values (\texttt{\#REF!}, \texttt{\#DIV/0!}, \texttt{\#NAME?}, \texttt{\#VALUE!}, circular references), and calculation mistakes. A high-quality financial model should produce accurate results and be free of technical errors that would undermine trust in the outputs. \textit{Evaluators assess}: Excel error values (\texttt{\#REF!}, \texttt{\#DIV/0!}, \texttt{\#NAME?}, \texttt{\#VALUE!}, \texttt{\#N/A}), circular reference warnings, broken or invalid cell references, logical errors in formulas, calculation mistakes, and inconsistent formulas across similar rows/columns.}
\centering
\small
\begin{tabular}{cp{12cm}}
\toprule
\textbf{Score} & \textbf{Anchor} \\
\midrule
1 & Multiple Excel errors present (\texttt{\#REF!}, \texttt{\#DIV/0!}, etc.), obvious calculation mistakes, circular references, or broken formulas that make the model unreliable\\
2 & Several errors or inaccuracies that need fixing; model produces questionable results \\
3 & Minor errors present but core calculations appear correct; needs cleanup but usable \\
4 & Very few errors; calculations are accurate with only trivial issues \\
5 & Error-free model; all formulas work correctly, calculations verified and accurate\\
\bottomrule
\end{tabular}
\end{table}

\begin{table}[h]
\caption{\textbf{Dimension 2: Formula Conventions.} \textit{Focus}: Separation of inputs from calculations; no hardcoded values in formulas. This criterion assesses whether the model follows best practices for formula construction. Inputs (assumptions, raw data) should be clearly separated from calculations. Formulas should reference input cells rather than containing hardcoded ``magic numbers.'' This makes models easier to audit, update, and understand. \textit{Evaluators assess}: Hardcoded numbers embedded in formulas (e.g., \texttt{=A1*0.35} instead of \texttt{=A1*\$B\$5}), clear input/assumption sections separate from calculations, use of cell references instead of typed values, the ``one row, one formula'' rule, consistent formula patterns across rows/columns, and ability to change assumptions with automatic propagation.}
\centering
\small
\begin{tabular}{cl}
\toprule
\textbf{Score} & \textbf{Anchor} \\
\midrule
1 & Hardcoded values throughout; no separation between inputs and calculations \\
2 & Many hardcoded values; inputs and calculations mixed together; difficult to audit \\
3 & Some separation of inputs; occasional hardcoded values; functional but not ideal \\
4 & Good separation of inputs from formulas; rare hardcoded values; easy to trace \\
5 & Exemplary separation; all assumptions in dedicated area; fully dynamic model \\
\bottomrule
\end{tabular}
\end{table}

\begin{table}[h]
\caption{\textbf{Dimension 3: Color Coding \& Visual Formatting.} \textit{Focus}: Professional, purposeful use of color and formatting. This criterion evaluates the visual presentation of the spreadsheet. Professional financial models use color purposefully-typically blue for inputs, black for formulas, green for links to other sheets, and optionally red for external links or data provider pulls. Excessive or inconsistent coloring (the ``rainbow effect'') is distracting and unprofessional. Good formatting enhances readability without being garish. \textit{Evaluators assess}: Consistent color scheme following finance conventions (blue for inputs/assumptions, black for formulas/calculations, green for cross-sheet links), absence of excessive ``rainbow'' formatting, professional font choices and sizes, consistent number formatting (decimals, percentages, currency), clear visual hierarchy, avoidance of merged cells, and clear distinction between headers/labels and data.}
\centering
\small
\begin{tabular}{cl}
\toprule
\textbf{Score} & \textbf{Anchor} \\
\midrule
1 & Garish ``rainbow'' formatting; colors obscure rather than clarify \\
2 & Excessive or random coloring; distracting visual noise \\
3 & Acceptable formatting; some color used but not consistently \\
4 & Good visual presentation; mostly consistent; professional with minor issues \\
5 & Clean, professional formatting; purposeful color coding; visually polished \\
\bottomrule
\end{tabular}
\end{table}

\begin{table}[h]
\caption{\textbf{Dimension 4: Structure \& Organization.} \textit{Focus}: Logical layout, clear sections, ease of audit. This criterion assesses how well the spreadsheet is organized for auditability. A well-structured model has a logical flow, clear sections, and is easy to navigate and audit. Information should be grouped sensibly, with inputs at the top or in a dedicated area, followed by calculations, and outputs clearly presented. \textit{Evaluators assess}: Logical top-to-bottom or left-to-right flow, clear section headers and labels, distinct Inputs/Workings/Outputs sections, grouping of related items, easy-to-follow calculation flow, navigation aids for multi-sheet models, and absence of scattered calculations in random cells.}
\centering
\small
\begin{tabular}{cl}
\toprule
\textbf{Score} & \textbf{Anchor} \\
\midrule
1 & Disorganized; calculations scattered randomly; very difficult to audit \\
2 & Poor organization; structure unclear; requires significant effort to follow \\
3 & Functional structure; can follow logic but organization could improve \\
4 & Well-organized; clear sections and flow; easy to navigate \\
5 & Excellent organization; intuitive layout; professional structure \\
\bottomrule
\end{tabular}
\end{table}

\begin{table}[h]
\caption{\textbf{Dimension 5: Financial Modeling Conventions.} \textit{Focus}: Adherence to standard financial modeling practices. This criterion evaluates whether the model follows established financial modeling conventions. This includes proper sign conventions, chronological time flow, integrity checks, and disciplined linking practices. A well-built model should be easy to audit without following complex reference chains. \textit{Evaluators assess}: Consistent sign convention (expenses uniformly negative or positive), chronological left-to-right time flow, checks and integrity tests (balance checks, control totals, error flags), linking discipline (direct links to source, no daisy-chaining), standard financial statement formats, proper treatment of beginning vs.\ ending balances, and avoidance of unnecessary circularity.}
\centering
\small
\begin{tabular}{cl}
\toprule
\textbf{Score} & \textbf{Anchor} \\
\midrule
1 & Ignores conventions; inconsistent sign treatment; would not pass professional review \\
2 & Multiple convention violations; difficult to reconcile with standard practices \\
3 & Mostly follows conventions with some inconsistencies; acceptable for draft work \\
4 & Good adherence to conventions; minor deviations; professional quality \\
5 & Exemplary adherence to financial modeling best practices throughout \\
\bottomrule
\end{tabular}
\end{table}

\begin{table}[h]
\caption{\textbf{Dimension 6: Purpose \& Practical Utility.} \textit{Focus}: Does the model accomplish its stated purpose? This criterion evaluates whether the spreadsheet actually accomplishes what the prompt asked for and presents outputs in a decision-useful way. Note: this is distinct from Errors \& Accuracy (which focuses on whether calculations are correct); here, focus on whether the model answers the prompt and is practically useful. \textit{Evaluators assess}: Whether the model addresses all parts of the prompt, presence of requested outputs/calculations, usefulness for actual decision-making, appropriate scope (neither missing key elements nor over-engineered), suitability for sharing with clients or stakeholders, clarity of results presentation, and provision of actionable insights.}
\centering
\small
\begin{tabular}{cl}
\toprule
\textbf{Score} & \textbf{Anchor} \\
\midrule
1 & Fails to address the prompt; missing key requirements; not useful \\
2 & Partially addresses prompt; significant gaps; limited practical utility \\
3 & Meets basic requirements; answers core question but lacks polish \\
4 & Good response to prompt; useful deliverable with minor gaps \\
5 & Fully addresses all aspects; excellent utility; ready for professional use \\
\bottomrule
\end{tabular}
\end{table}

\subsection{Overall Rating}

The overall rating is computed as the arithmetic mean of the six dimension scores, rounded to the nearest integer:
\begin{align}
    \mathrm{Overall} = \mathrm{round}\left(\frac{1}{6}\sum_{i=1}^{6} C_i\right)
\end{align}
where $C_i$ denotes the score for dimension $i$.

\subsection{Circulation Question}
\label{subsec:circulation}
In addition to the six dimension scores, evaluators answered a single binary question for each spreadsheet: \textit{``Would you share this with a client or colleague?''} (yes/no). This question captures a holistic, practitioner-level judgment of professional acceptability that complements the per-dimension ratings: a spreadsheet may score acceptably on individual criteria yet still fall short of the bar a practitioner would apply before circulating it.

\subsection{Detailed Results}
Beyond the headline means, functional criteria were strongest: \emph{Errors \& Accuracy} ($M=3.43$, 75.6\% $\geq 3$) and \emph{Formula Conventions} ($M=3.10$, 70.4\% $\geq 3$). Adherence was weaker for \emph{Modeling Conventions} ($M=2.68$, 40.4\% $\leq 2$) and \emph{Purpose \& Utility} ($M=2.58$, 47.2\% $\leq 2$). The largest deficiency was \emph{Color Coding and Formatting} ($M=1.97$, $\mathrm{SD}=0.77$), with 79.6\% scoring $\leq 2$ and only 4.0\% scoring $\geq 4$. Experts indicated they would circulate only 16.1\% of spreadsheets to a client or colleague.

\subsection{Alignment and Reliability}
Across 25 battles, expert ratings agreed with arena outcomes in 56.0\% of cases, disagreed in 32.0\%, and tied in 12.0\%. Among decisive comparisons, agreement was 63.6\%, moderately above chance. Across all 50 spreadsheets (each rated by all five experts), Krippendorff's $\alpha$ ranged from 0.28 to 0.45 across dimensions, indicating low inter-rater reliability. Despite variability in precise rankings, aggregate scores suggest only partial adherence to professional financial standards.

\section{Efficiency Analysis}
\label{app:efficiency-analysis}

Table~\ref{tab:token-usage} shows per-model token usage. Table~\ref{tab:api-cost} shows per-model cost at published API rates.

\begin{table}[t]
\caption{Per-model token usage. Token counts reflect total compute used, including models that were executed but may not have been voted on.}
\label{tab:token-usage}
\centering
\footnotesize
\setlength{\tabcolsep}{4pt}
\begin{tabular}{lrrrrrr}
\toprule
Model & Execs & Avg In & Avg Out & Total In & Total Out & Total \\
\midrule
Claude Sonnet 4.5 & 1{,}003 & 1{,}650 & 11{,}438 & 1{,}655{,}361 & 11{,}472{,}684 & 13{,}128{,}045 \\
Claude Opus 4.5   &    749 & 2{,}006 & 12{,}030 & 1{,}502{,}598 &  9{,}010{,}254 & 10{,}512{,}852 \\
Claude Opus 4.1   & 1{,}006 & 1{,}676 &  6{,}425 & 1{,}686{,}376 &  6{,}463{,}990 &  8{,}150{,}366 \\
GPT-5             &    751 & 1{,}192 &  9{,}389 &    895{,}099 &  7{,}051{,}064 &  7{,}946{,}163 \\
Gemini 2.5 Pro    &    991 &    310 &  6{,}693 &    307{,}502 &  6{,}632{,}289 &  6{,}939{,}791 \\
GPT-5.2           &    581 & 1{,}952 &  8{,}128 & 1{,}133{,}957 &  4{,}722{,}612 &  5{,}856{,}569 \\
Gemini 3 Pro      &    777 &    337 &  5{,}630 &    261{,}938 &  4{,}374{,}301 &  4{,}636{,}239 \\
GPT-5.1           &    734 & 1{,}535 &  4{,}428 & 1{,}126{,}466 &  3{,}249{,}957 &  4{,}376{,}423 \\
Grok Code Fast 1  &    679 & 1{,}727 &  3{,}596 & 1{,}172{,}508 &  2{,}441{,}965 &  3{,}614{,}473 \\
Kimi K2 Instruct  &    276 & 5{,}833 &  2{,}896 & 1{,}609{,}781 &    799{,}410 &  2{,}409{,}191 \\
GPT-4o            &    837 & 1{,}501 &  1{,}367 & 1{,}256{,}185 &  1{,}143{,}822 &  2{,}400{,}007 \\
Grok 4.1 Fast     &    534 & 1{,}870 &  2{,}526 &    998{,}663 &  1{,}348{,}620 &  2{,}347{,}283 \\
Qwen3 30B         &    151 & 7{,}459 &  6{,}475 & 1{,}126{,}327 &    977{,}764 &  2{,}104{,}091 \\
Llama 4 Maverick  &    176 & 8{,}205 &  2{,}120 & 1{,}444{,}139 &    373{,}166 &  1{,}817{,}305 \\
Gemini 2.5 Flash  &    220 &    223 &  4{,}971 &     49{,}070 &  1{,}093{,}679 &  1{,}142{,}749 \\
Grok 4            &    182 & 1{,}574 &  1{,}281 &    286{,}477 &    233{,}151 &    519{,}628 \\
\midrule
\textbf{Total} & \textbf{9{,}647} & \textbf{1{,}712} & \textbf{6{,}363} & \textbf{16{,}512{,}447} & \textbf{61{,}388{,}728} & \textbf{77{,}901{,}175} \\
\bottomrule
\end{tabular}
\end{table}

\begin{table}[t]
\caption{API costs by model using published provider pricing (\$/1M tokens, input/output). Total experimental cost for the frozen snapshot is \$1{,}244.10 across 9{,}647 generations. Anthropic models account for 74\% of total cost, driven primarily by Claude Opus pricing.}
\label{tab:api-cost}
\centering
\footnotesize
\setlength{\tabcolsep}{4pt}
\begin{tabular}{llrrrlr}
\toprule
Model & Provider & Execs & Input Tok & Output Tok & Rate (in/out) & Cost \\
\midrule
Claude Opus 4.1   & anthropic & 1{,}006 & 1{,}686{,}376 &  6{,}463{,}990 & \$15.0 / 75.0 & \$510.09 \\
Claude Opus 4.5   & anthropic &    749 & 1{,}502{,}598 &  9{,}010{,}254 &  \$5.0 / 25.0 & \$232.77 \\
Claude Sonnet 4.5 & anthropic & 1{,}003 & 1{,}655{,}361 & 11{,}472{,}684 &  \$3.0 / 15.0 & \$177.06 \\
GPT-5             & openai    &    751 &    895{,}099 &  7{,}051{,}064 & \$1.25 / 10.0 &  \$71.63 \\
GPT-5.2           & openai    &    581 & 1{,}133{,}957 &  4{,}722{,}612 & \$1.75 / 14.0 &  \$68.10 \\
Gemini 2.5 Pro    & gemini    &    991 &    307{,}502 &  6{,}632{,}289 & \$1.25 / 10.0 &  \$66.71 \\
Gemini 3 Pro      & gemini    &    777 &    261{,}938 &  4{,}374{,}301 &  \$2.0 / 12.0 &  \$53.02 \\
GPT-5.1           & openai    &    734 & 1{,}126{,}466 &  3{,}249{,}957 & \$1.25 / 10.0 &  \$33.91 \\
GPT-4o            & openai    &    837 & 1{,}256{,}185 &  1{,}143{,}822 &  \$2.5 / 10.0 &  \$14.58 \\
Grok 4            & xai       &    182 &    286{,}477 &    233{,}151 &  \$3.0 / 15.0 &   \$4.36 \\
Grok Code Fast 1  & xai       &    679 & 1{,}172{,}508 &  2{,}441{,}965 &  \$0.2 / 1.5  &   \$3.90 \\
Kimi K2 Instruct  & fireworks &    276 & 1{,}609{,}781 &    799{,}410 &  \$0.6 / 2.5  &   \$2.96 \\
Gemini 2.5 Flash  & gemini    &    220 &     49{,}070 &  1{,}093{,}679 &  \$0.3 / 2.5  &   \$2.75 \\
Grok 4.1 Fast     & xai       &    534 &    998{,}663 &  1{,}348{,}620 &  \$0.2 / 0.5  &   \$0.87 \\
Qwen3 30B         & fireworks &    151 & 1{,}126{,}327 &    977{,}764 & \$0.15 / 0.6  &   \$0.76 \\
Llama 4 Maverick  & fireworks &    176 & 1{,}444{,}139 &    373{,}166 & \$0.22 / 0.88 &   \$0.65 \\
\midrule
\textbf{Total} & & \textbf{9{,}647} & \textbf{16{,}512{,}447} & \textbf{61{,}388{,}728} & & \textbf{\$1{,}244.10} \\
\bottomrule
\end{tabular}
\vspace{4pt}
\\

\end{table}

\end{document}